\pdfoutput=1
\documentclass{article}

\usepackage{microtype}
\usepackage{graphicx}
\usepackage{subfigure}
\usepackage{booktabs} 

\usepackage[accepted]{icml2020}
\usepackage{xr-hyper}
\usepackage{hyperref}

\usepackage[utf8]{inputenc}
\usepackage[T1]{fontenc}

\usepackage[accepted]{icml2020}
\icmltitlerunning{Emergence of Separable Manifolds in Deep Language Representations}

\begin{document}
\setlength{\parskip}{0.5em}

\twocolumn[
\icmltitle{Emergence of Separable Manifolds in Deep Language Representations}



\icmlsetsymbol{equal}{*}

\begin{icmlauthorlist}
\icmlauthor{Jonathan Mamou}{equal,intel}
\icmlauthor{Hang Le}{equal,mit}
\icmlauthor{Miguel A Del Rio}{mit}
\icmlauthor{Cory Stephenson}{intel}

\icmlauthor{Hanlin Tang}{intel}
\icmlauthor{Yoon Kim}{hv}
\icmlauthor{SueYeon Chung}{mit,col}
\end{icmlauthorlist}

\icmlaffiliation{intel}{Intel Labs}
\icmlaffiliation{mit}{Massachusetts Institute of Technology}
\icmlaffiliation{hv}{Harvard University}
\icmlaffiliation{col}{Columbia University}

\icmlcorrespondingauthor{SueYeon Chung}{sueyeon@mit.edu}

\icmlkeywords{Machine Learning, ICML}

\vskip 0.3in
]



\printAffiliationsAndNotice{\icmlEqualContribution} 

\begin{abstract}
Deep neural networks (DNNs) have shown much empirical success in solving perceptual tasks across various cognitive modalities. While they are only loosely inspired by the biological brain, recent studies report considerable similarities between representations extracted from task-optimized DNNs and neural populations in the brain. DNNs have subsequently become a popular model class to infer computational principles underlying complex cognitive functions, and in turn, they have also emerged as a natural testbed for applying methods originally developed to probe information in neural populations. In this work, we utilize mean-field theoretic manifold analysis, a recent technique from computational neuroscience that connects geometry of feature representations with linear separability of classes, to analyze language representations from large-scale contextual embedding models. We explore representations from different model families (BERT, RoBERTa, GPT, etc.) and find evidence for emergence of linguistic manifolds across layer depth (e.g.,
manifolds for part-of-speech tags), especially in ambiguous data (i.e, words with multiple part-of-speech tags, or part-of-speech classes including many words). In addition, we find that the emergence of linear separability in these manifolds is driven by a combined reduction of manifolds’ radius, dimensionality and inter-manifold correlations.\footnote{Code is available at \url{https://github.com/schung039/contextual-repr-manifolds}.} 
\end{abstract}

\section{Introduction and Related Work}
Many recent studies show notable similarities between representations extracted from task-optimized deep neural networks (DNNs) and neural populations in the brain in sensory systems \cite{yamins2014performance, khaligh2014deep}. Computational neuroscience community is increasingly relying on utilizing DNNs as a framework for studying neural correlates underlying complex cognitive functions \cite{cichy2019deep, kriegeskorte2015deep}. Addressing the question of "how a population of neural units transform representations across multilayered processing stages to implement a cognitive task" is a key challenge in both neuroscience and deep learning. Consequently, developing the techniques to provide insight into neural representation and computation have been an active area of research in both fields \cite{barrett2019analyzing}. 

Much prior work on characterizing how information is encoded in DNNs and the brain has focused on the geometric structure underlying the data. In neuroscience, representational similarity analysis \cite{kriegeskorte2013representational} captures the similarity between the stimuli in the geometry of the neural data and deep network representations. Other geometric measures such as geodesics \cite{henaff2015geodesics}, curvature \cite{henaff2019perceptual, fawzi2018empirical}, intrinsic dimension  \cite{ansuini2019intrinsic}, and canonical correlation analysis \cite{raghu2017svcca} have been used to empirically study the complexity of neural population and learned representations in DNNs.

In natural language processing (NLP), recent advances in contextualized word representations such as ELMo \cite{peters-etal-2018-deep} and BERT  \cite{bert} have led to significant empirical improvements across many tasks. Concomitant with these advances is an emergent line of work, colorfully referred to as \emph{BERTology}, exploring what aspects of language are being captured by these contextual representations~\cite{zhang2018language,blevins2018deep,tenney2019bert}. One popular approach for analysis is also through the lens of the geometry: \citet{stanford_nlp} report evidence of a geometric representation of parse trees in embedding from BERT, and \citet{coenen2019} study the geometric representation of word senses via visualization techniques such as UMAP. Another popular approach for analyzing these representations is through \emph{supervised probes}, i.e. classifiers  trained on top of fixed representations to predict certain linguistic properties (e.g., part-of-speech tags, syntactic heads)~\cite{liu_et_al,tenney2019you}.
Supervised probes are conceptually simple and have greatly expanded our understanding of the kinds of linguistic knowledge encoded by these models. However, they are unable to capture the intrinsic geometry underlying the learned representation space, and it is not clear that high accuracy with respect to a probing task necessarily implies that the relevant linguistic structure is being encoded.

In this paper, we apply a recent manifold analysis technique based on replica theory ~\cite{chung2018classification} that links the geometry of object manifolds to the shattering capacity of a linear classifier as a measure of the amount of information stored about object categories per unit. This method has been used in sensory domains such as visual CNNs~\cite{cohen2019separability}, visual neuroscience~\cite{chung2020separable} and deep speech recognition models~\cite{stephenson2019untangling} to characterize how object manifolds ‘untangle’ across layers. Here we apply this manifold analysis to study deep language representations, particularly Transformer-based models \cite{vaswani2017}, for the first time, and show that NLP systems also "untangle" linguistic ``objects'' relevant for the task. 

We present several key findings:
\begin{enumerate}
\item Word and linguistic category manifolds emerge across the deep layers of Transformer architectures, in the task-dependent, predictive regime (where the feature vectors are defined on masked tokens), similar to vision and speech deep networks. 
\item In word contextualization regime (defined on unmasked tokens), word manifolds strongly decrease in the manifold capacity, becoming less separable across the hierarchy. Linguistic manifolds are affected by the underlying word manifolds, but are counteracted by the contextualization, resulting in linguistic manifolds with a better effective separation compared to word manifolds.
\item The emergence of part-of-speech (POS) manifolds is observed most strongly when the underlying words are ambiguous with multiple POS tags in BERT. POS manifolds further seem to interpolate between word-like geometry and separable contextual geometry, depending on the number of words in each POS class. 
\end{enumerate}

In addition, we show the generality of linguistic untangling with word representation manifolds in widely-utilized NLP models. We also show that geometry of fine-tuning learning dynamics can be probed with the tasks congruent, incongruent to the training, to measure the similarity between the tasks. These results provide geometric evidence for emergence of language representation manifolds, from words to parts-of-speech to named entities, in deep neural networks for natural language processing.

\section{Mean-Field Theoretic Manifold Analysis}
\label{sec:MFT_analysis_intro}
\begin{figure}[h]
\begin{center}
\centerline{\includegraphics[width=\columnwidth]{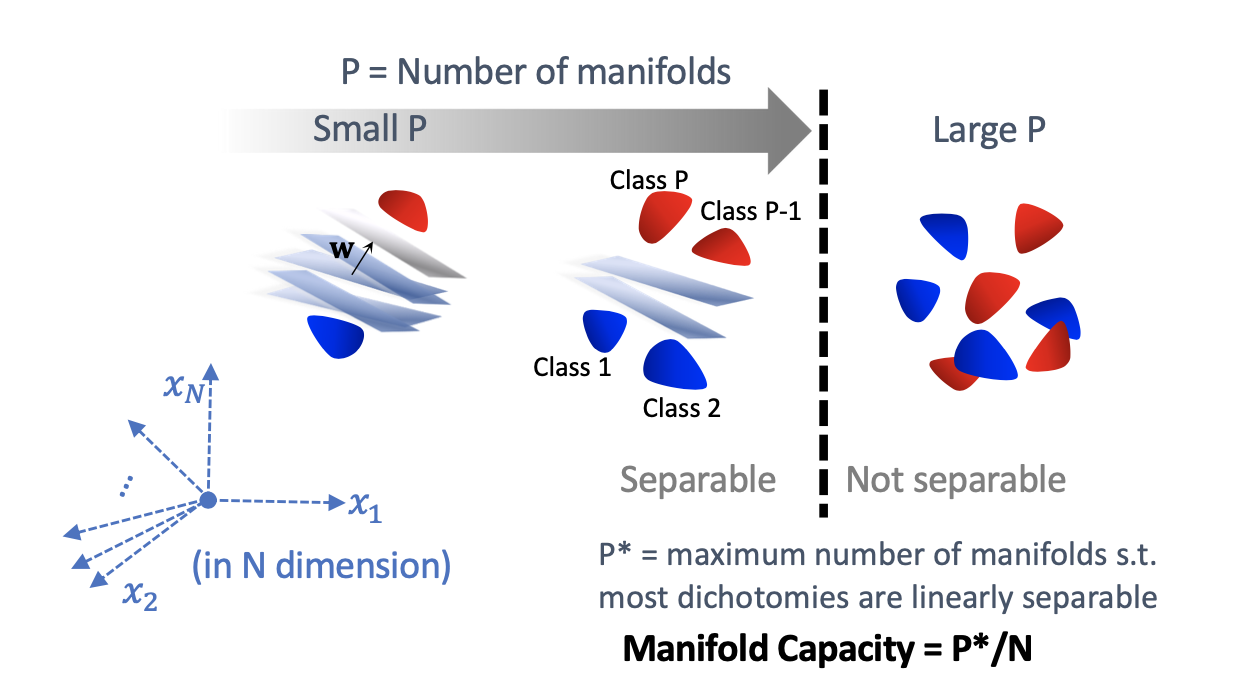}}
\vspace{-3mm}
\caption{\textbf{Manifold Capacity} measures the linear separability of class manifolds, and it is a generalization of perceptron capacity to classes.}
\vspace{-6mm}
\label{fig:capacity_illu}
\end{center}
\end{figure}
\begin{figure*}[h]
\begin{center}
\centerline{\includegraphics[width=\textwidth]{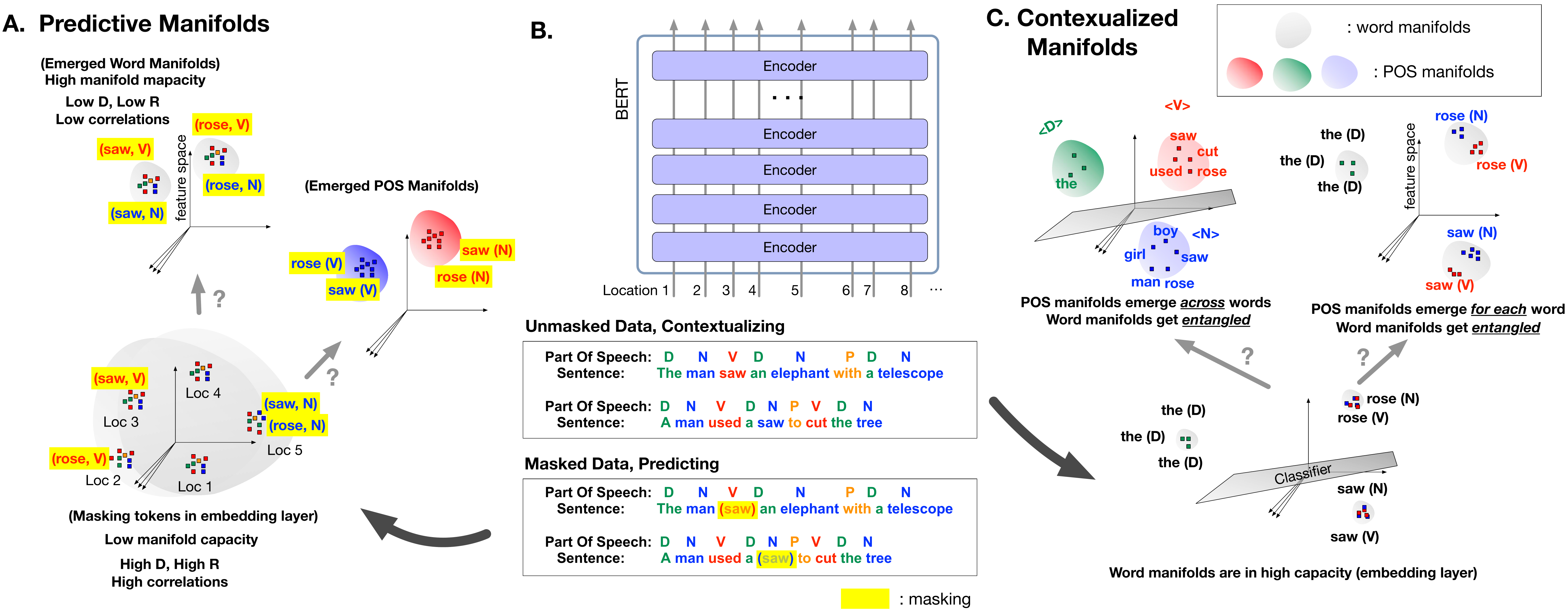}}
\vspace{-3mm}
\caption{Illustration of a hypothesis that word and linguistic manifolds emerge across a Transformer hierarchy: (A) Predictive manifolds defined on masked tokens. (B) BERT hierarchy. (C) Contextualized manifolds defined on unmasked tokens.}
\vspace{-4.5mm}
\label{fig:intro_mft_lang}
\end{center}
\end{figure*}

This paper uses mean-field theoretic manifold analysis~\cite{chung2018classification,stephenson2019untangling} (hereafter, MFTMA technique) as a core analytic tool to measure manifold capacity and other manifold geometric properties (radius, dimension, center correlation) on a subsample of the test dataset. 

Given $P$ object manifolds (i.e. feature vectors with their categories) in $N$ feature dimensions, \textit{Manifold capacity} (Fig. \ref{fig:capacity_illu}), $\alpha_{M}=P/N$, refers to the critical number of object manifolds, $P$, that can be linearly separated given $N$ features.  $\alpha_M$ marks the value above which most manifold dichotomies are inseparable, and below which most are separable with a linear classifier. This is similar to the perceptron capacity for discrete points \cite{Gardner1988}, except that the unit for the numerator in the perceptron capacity is the number of category manifolds, rather than the number of discrete patterns. The manifold capacity thus measures the linearly separable information about object identity per feature. The manifold capacity can be computed empirically (\textit{Simulation manifold capacity}, $\alpha_{SIM}$, hereafter), by doing a bisection search to find the critical number of features N such that the fraction of linearly separable random manifold dichotomies is close to 1/2. MFTMA framework theoretically predicts manifold capacity, and connects it with the geometric properties of category manifolds. As such, MFTMA framework returns four quantities below: 

\begin{enumerate}
    \item \textbf{Mean Field Theoretic (MFT) Manifold Capacity ($\alpha_M$)} estimates the manifold capacity defined above, using the replica mean field formalism introduced by \citet{chung2018classification}.
    
    \item \textbf{Manifold Dimension ($D_M$)} captures the dimensions of an object manifold, and estimates the average embedding dimension of the examples contributing to the decision hyperplane. \par

    \item \textbf{Manifold Radius ($R_M$)} captures the size of the manifold relevant for linear separability, relative to the norm of the manifold centroid. Small manifold radius implies that the set of examples that determine the decision boundary are tightly grouped. \par
    
    \item \textbf{Center Correlations ($\rho_{center}$)}  measures the average of absolute values of the pairwise correlations between manifold centroids.

\end{enumerate}

Note that simulation manifold capacity has been reported to be accurately predicted by MFT manifold capacity. We provide their consistency in our data in Fig. \ref{fig:capacity_bert_ntasks}. In this paper, we use MFT Manifold Capacity (hereafter, Manifold Capacity, $\alpha_{M}$). The lower bound of the manifold capacity from data is given by \citet{cover1965geometrical}, and reflects the case where there is no manifold structure. Manifold capacity in most random initialized DNNs closely follow lower bound capacity, and we find similar trends in language models (see SM~\ref{SI:random}).

The key feature of MFTMA is that manifold capacity can be predicted by the geometric properties of the object manifolds, i.e., Manifold Dimension, Manifold Radius, and their center correlations. Small values for manifold radius, dimension and center correlation result in larger manifold capacity, rendering a more favorable geometry for classification.
\par 
\vspace{-3mm}

\section{Experimental Setup}
\vspace{-1mm}
We apply MFTMA to study the geometry of representations from a variety of contextualized word embedding models. We target Transformer-based models \cite{vaswani2017}, which compose contextual representations at each layer with self-attention, as they have been shown to produce state-of-the-art results across a large number of NLP tasks. Transformer networks also provide an opportunity to analyze the evolution of representations across layer depth since they typically employ more hidden layers than other neural NLP models.
\vspace{-1mm}
\subsection{Models}
\vspace{-1mm}

\paragraph{BERT-based architectures} {\bf BERT}~\cite{bert} is a bidirectional Transformer pre-trained using a combination of masked language modeling objective and next sentence prediction on a large corpus.
We also analyzed different architectures derived from BERT:  {\bf RoBERTa}~\cite{liu2019roberta} modifies key hyperparameters in BERT including removing the next-sentence pre-training objective, and training with much larger mini-batches and learning rates; {\bf ALBERT}~\cite{lan2019albert} uses parameter-reduction techniques to lower memory consumption and increase the training speed of BERT; {\bf DistilBERT}~\cite{sanh2019distilbert} is a  Transformer model trained by distilling BERT into a smaller model. 
\vspace{-2mm}
\paragraph{GPT architecture} {\bf OpenAI GPT}~\cite{radford2018improving} is a unidirectional left-to-right Transformer pre-trained using language modeling on a large corpus.

\vspace{-2mm}
\paragraph{Pretrained models} We use the following model versions pretrained on English text: {\tt bert-base-cased}, {\tt albert-base-v1}, {\tt roberta-base}, {\tt distilbert-base-uncased}, and {\tt openai-gpt}. All the pre-trained models use a 12-layer transformer except {\tt distilbert-base-uncased} that uses a 6-layer transformer and have hidden size of 768.
\vspace{-2mm}
\paragraph{Fine-tuned models} In order to analyze how representations change when the model parameters are optimized towards a different task, we also test our approach on models fine-tuned on a downstream task (POS) for different model training {\it updates steps}.

\vspace{-1mm}

\subsection{Datasets and Manifold Definitions}
\vspace{-1mm}
\label{ssec:datasets}

As noted in section~\ref{sec:MFT_analysis_intro}, MFTMA begins by assigning each representation to a particular linguistic category (i.e. manifold). We 
 experiment with various word-level categories (derived from common NLP tasks/datasets) that target different language phenomena. 
\vspace{-2mm}
\paragraph{Word} A word manifold  (\texttt{word}) contains several instances of the same word occurring in different contexts. We use the Penn Treebank (PTB) ~\cite{marcus-etal-1993-building} and select 80 word manifolds based on most frequent words in the corpus.
\vspace{-2mm}
\paragraph{Part-of-speech tags} POS (\texttt{pos}) tags consists of tags such as \textit{proper nouns} (\textit{NNP}), \textit{determiners} (\textit{DT}), etc., and are typically considered to target lower-level syntactic phenomena. We select the 33 most frequent tags from PTB.

\vspace{-2mm}
\paragraph{Semantic tags} We use the semantic tagging (\texttt{sem-tag}) dataset by \citet{abzianidze2017towards}, which annotates words with semantically informative tags. We  take the 61 most-frequent semantic tags. Some examples of these tags include {\it comparative positive}, {\it concept}, {\it implication}, etc (we refer the reader to the original paper for the full tag definitions). This dataset has also been utilized to analyze contextual word representations in the context of supervised linear probes \cite{liu_et_al}. 
\vspace{-2mm}
\paragraph{Named-entity recognition} This task (\texttt{ner}) consists of locating and classifying named entity mentioned in text into pre-defined categories such as {\it person names}, {\it organizations}, {\it locations}, etc. It allows for finer-grained manifolds than other tasks since it involves additional segmentation
using BIO (\textit{begin}, \textit{inside}, \textit{outside}) tags. We use the tags from the Ontonotes dataset~\cite{weischedel2011ontonotes}.
\vspace{-2mm}
\paragraph{Dependency depth} All of the above datasets/tags (except words) have been studied by \citet{liu_et_al}. Following \cite{hewitt2019}, we study representations stratified by depth in a dependency tree (\texttt{dep-depth}). For each contextualized word representation, we use its depth in a dependency tree to assign it to a depth manifold. We select the 22 most frequent depths from PTB.

For each of the word-level tags defined above, we randomly sample 50 word instances per tag to perform the manifold analysis. We average the manifold metrics across five repetitions.
For the rest of the paper, we use {\it linguistic manifolds}  to refer to manifolds based on \texttt{word}, \texttt{pos}, \texttt{sem-tag}, \texttt{ner} and \texttt{dep-depth}. We further use {\it linguistic category manifolds} to refer to all linguistic manifolds except for dependency depth, since depth in a dependency tree is numeric.

\vspace{-1mm}
\subsection{Feature Extraction}
\vspace{-1mm}
BERT-based models are trained with a masked language modeling objective
which randomly replaces words in a sentence with a special \texttt{[MASK]} token. We experiment with two encoding schemes for obtaining the contextualized representations in BERT-based models: \emph{masked} and \emph{unmasked}. In the \emph{masked} case, we use the \texttt{[MASK]} token to obtain the contextualized representation, and assign the representation to the linguistic category of the predicted word (\emph{predictive manifold} in Fig.~\ref{fig:intro_mft_lang}). In the \emph{unmasked} case, we obtain the contextualized word embedding by using the actual token as the input, correspondingly assign the representation to the linguistic category of the input (\emph{contextualized manifold} in Fig.~\ref{fig:intro_mft_lang}). In practice we observe that these two encoding schemes lead to differences in whether linguistic manifolds emerge in earlier or later layers of the network.\footnote{\citet{voita2019} also analyze representations stratified by masked/unmasked representations, and report considerable differences the evolution of mutual information across layers.}
If a word is tokenized into multiple tokens (subwords), its word representation is computed as the average of all its subwords' representations.

\subsection{Methods}

Our mean field theoretic manifold analysis closely follows prior work by \citet{stephenson2019untangling}. We supplement the manifold analysis with two additional techniques.

\paragraph{Distribution of SVM Fields}
\label{sssection:method_SVMfields}
We analyze the distribution of fields (i.e. margins), defined as the signed perpendicular distance between the feature vectors and the optimal linearly separating hyperplane using support vector machines (SVM). We train a slack-SVM with a linear kernel. Given $P$ classes (e.g. $P$ POS tags), we train SVM classifiers for one-versus-rest classification task. The fields are measured only for the feature vectors with positive ground truth label (i.e. "one" of one vs. rest classification). For a given class, the fields from the true positives and false negatives are collected, and normalized by the field distance between the positive and negative class centroids. We collect these normalized fields from all P classes to finally obtain the distribution over fields. We also provide the similar analysis without the normalization (true perpendicular distances to the hyperplane) for reference in SM~\ref{SI:SVM}. The tail of the distribution on the positive side reflects a linear classification accuracy. 
\vspace{-2mm}
\begin{figure}[h]
\begin{center}
\centerline{\includegraphics[width=0.8\columnwidth]{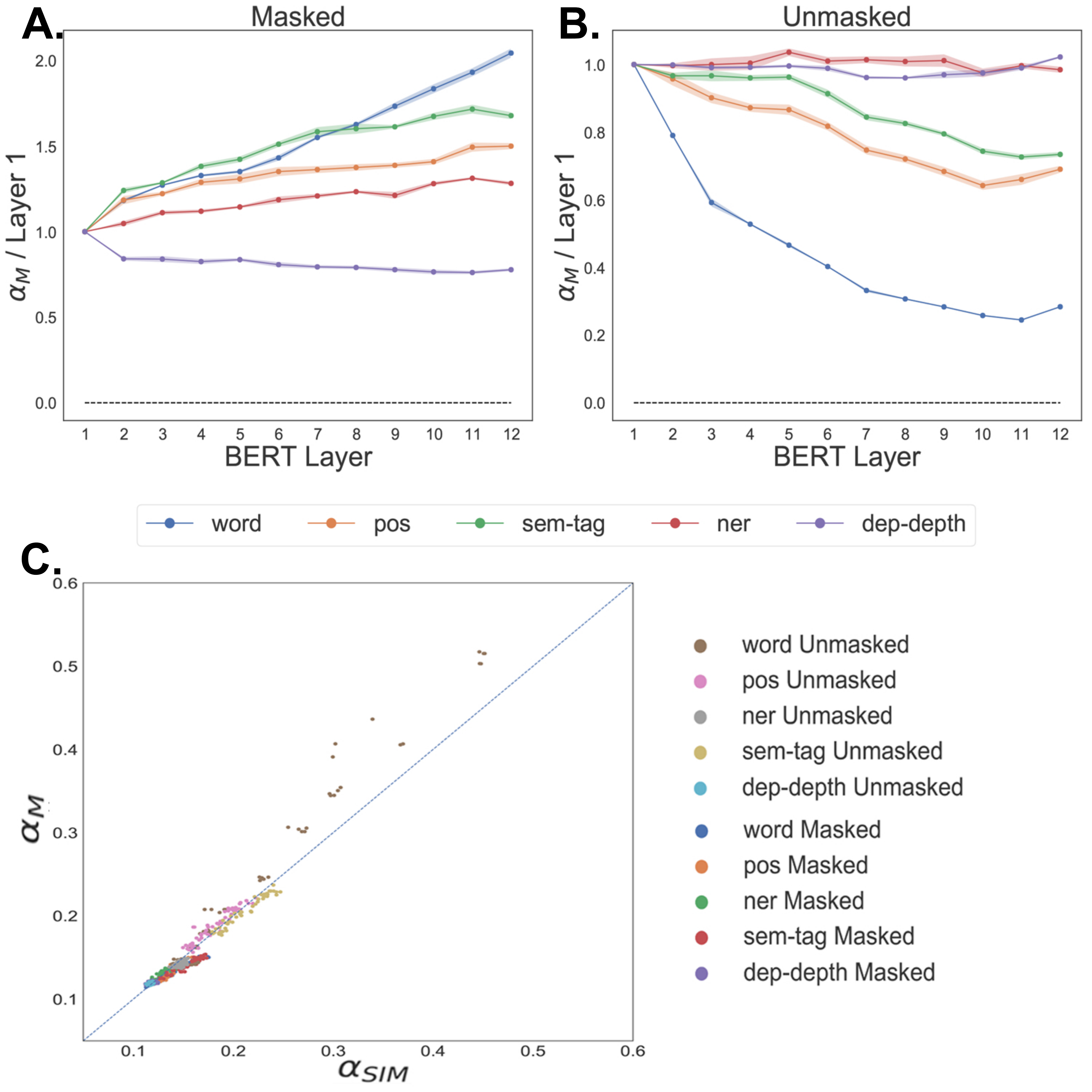}}\vspace{-3mm}
\caption{\textbf{Manifold capacity emerges across various linguistic tags in BERT } Manifolds defined by tags for words, POS, sem-tag, NER, dep-depth from (A) masked and (B) unmasked data (Note that empirical capacity matches MFT theory (C)). In A-B, MFT manifold capacity is normalized by manifold capacity of the first layer for each tags ("Capacity/First layer"), to compare the effect across different tasks. Error bars hereafter represent standard error on the mean of the calculated manifold metrics (including 5 random seeds)} \vspace{-8mm} \label{fig:capacity_bert_ntasks}
\end{center}
\end{figure}
\paragraph{PCA visualizations}
\label{sssection:method_PCA} To qualitatively analyze the evolution of representations across layers, we visualize the representations with PCA, where each data point is color-coded according to its tag. When comparing representations across multiple layers, we perform PCA on data across all layers. 

\vspace{-4mm}
\section{Results}
\subsection{Emergence of Separable Language Manifolds}
\label{ssection:separableManifolds}
\paragraph{MFTMA analysis}
\label{sssection:MFTMA}
We first investigated the manifold capacity of language manifolds in the BERT model, which was trained to predict the word identity of the masked tokens in the output layer, using the datasets described in \ref{ssec:datasets}. The datasets have two forms: ``Masked", to observe the emergent properties of the task-dependent, predictive language manifolds, and ``Unmasked", to probe the information content in the contextualized word manifolds.\footnote{Note that BERT is trained on masked tokens,  but uses unmasked tokens at test time.} The manifold capacity is presented as the relative change compared to capacity of the first layer (Fig.~\ref{fig:capacity_bert_ntasks}, A and B) to enable comparison of the linguistic content embedded in the representations between different datasets. 

\begin{figure}[h]
\begin{center}
\centerline{\includegraphics[width=\columnwidth]{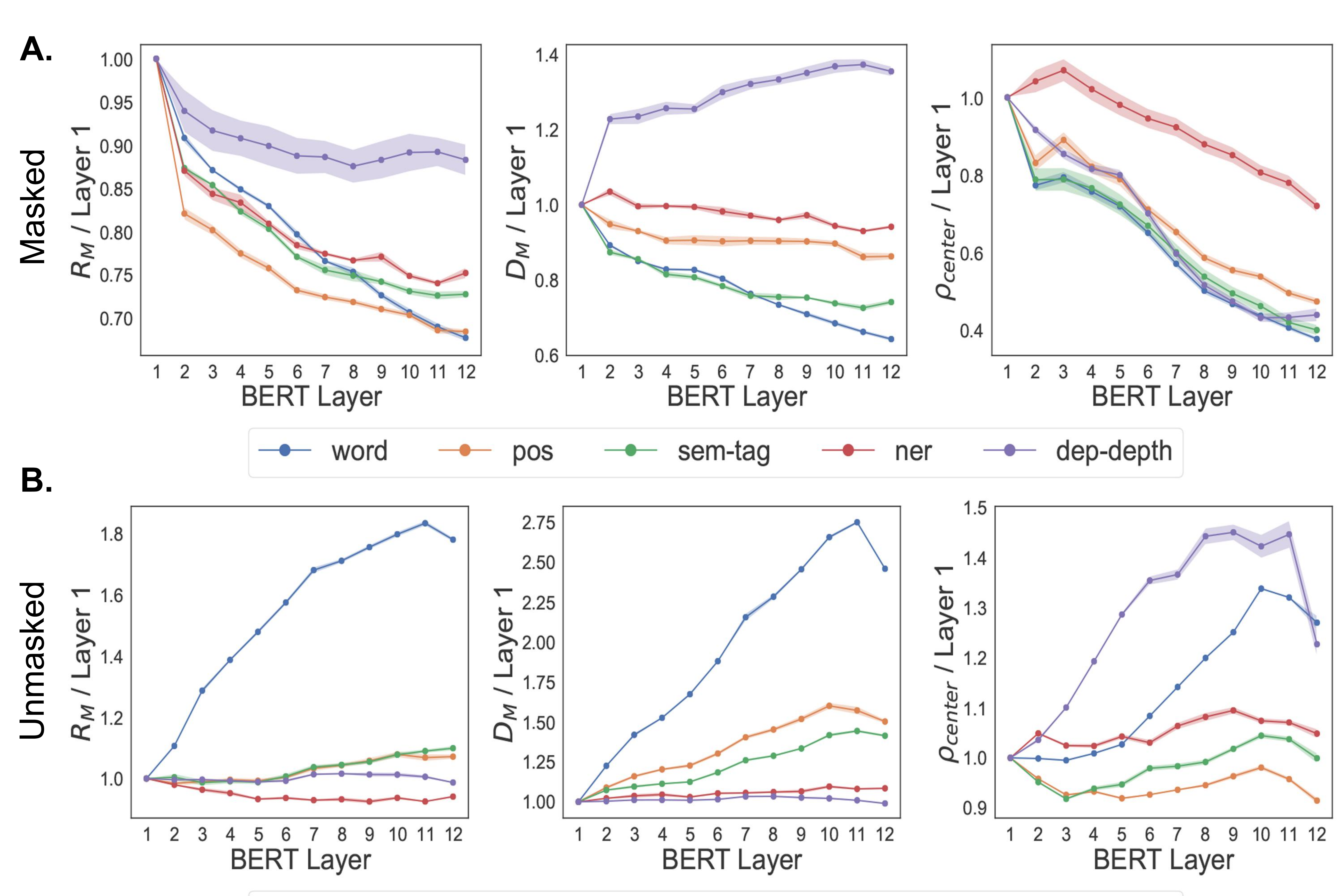}}
\vspace{-2mm}
\caption{\textbf{Manifold Dimension, Radius and Center Correlations for various tasks in BERT} Geometrical property for different linguistic manifolds: words, POS, sem-tag, ner, dep-depth from masked (A) and unmasked (B) data. These geometrical properties are normalized by the property of the first layer.} \vspace{-7mm}
\label{fig:mft_geometry_bert_ntasks}
\end{center}
\end{figure}
Since the model is explicitly trained with masked modeling task on the word level, we observe that the word and other linguistic category manifold classes become more separable and increase in their capacity, over the course of the layers, on the masked data (Fig. \ref{fig:capacity_bert_ntasks}A). Interestingly, other language classes based on POS, ner, sem-tag categories ("linguistic category manifolds") also emerges across the layers, and surprisingly, their relative increase across layers is comparable to the increase in word manifolds, perhaps reflecting the fact that the emergent linguistic category information is used to predict the masked words in the final layer. 

On the unmasked data, since the word token is present in the input signal and is dominant, word capacity is much higher than the capacity in the masked data (Fig. \ref{fig:capacity_bert_ntasks}B). As the input word embedding is already well separated, the word separability is at its highest in the embedding layer as expected. In subsequent layers, these highly separable word manifolds become contextualized and their capacity significantly decreases, as shown in Fig. \ref{fig:capacity_bert_ntasks}B. For other language manifolds such as POS or ner, where the underlying representations are also based on word features (except with language tags to define partitions based on linguistic categories), the capacity is similarly diminished, although not to the degree compared to the words manifold, due to the effect of the contextualization (Fig.~\ref{fig:mft_geometry_bert_ntasks}). Unlike the masked case, the linguistic manifolds have generally smaller capacity compared to words (Fig. \ref{fig:capacity_bert_ntasks}B). 

While the manifold capacity indicates an emergence of separable manifolds, the mean-field geometric metrics such as the manifold dimension, radius, and correlations tells us "how" the separability arises, i.e., theoretically-grounded geometric evidence of untangling. Across multiple categories (POS, NER, sem-tag, etc.), we find that for the masked data, the increased word and linguistic manifolds capacities are due to the reduction in the manifold radius, dimension, and center correlations (Fig. \ref{fig:mft_geometry_bert_ntasks}A), similar to prior work in vision and speech \cite{stephenson2019untangling}. For the unmasked case, the opposite trends are observed (Fig. \ref{fig:mft_geometry_bert_ntasks}B). 

\begin{figure*}[h]
\begin{center}
\centerline{\includegraphics[width=1.5\columnwidth]{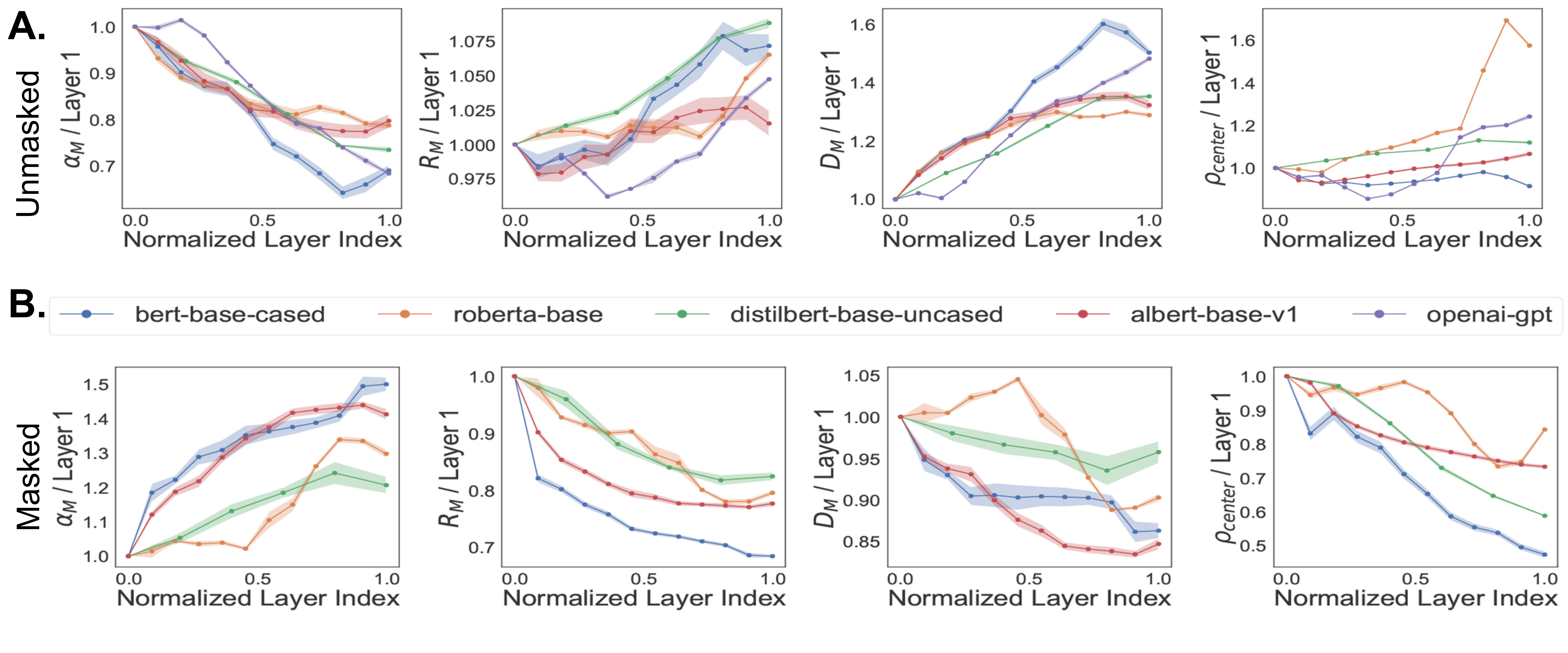}}
\vspace{-5mm}
\caption{\textbf{Emergence of language manifolds across different transformer classes, shown by POS manifolds} The evolution of manifold capacity, manifold dimension, manifold radius and center correlations within different models (BERT, roBERTa, alBERT, distilBERT and openAI-GPT) for unmasked data (A) and masked data (B). All values are normalized by the first layer. The layer index is normalized so that different models can be plotted together.} \vspace{-10mm}
\label{fig:mft_manymodels}
\end{center}
\end{figure*}

We also find that different linguistic categories show different relative trends. For manifold capacities (Fig. \ref{fig:capacity_bert_ntasks}), in masked data, words, POS, sem-tag, ner show comparable amount of emergent capacity, while dep-depth shows a general reduction across layers. This might be due to the fact that dep-depth is based on depth of the tree, unlike other category-based tags, and classification-based metrics might not capture their functional and structural emergence. In unmasked data, a large decrease in the manifold capacity of POS (as compared to ner) potentially indicates that much POS tag information can be derived from words alone, which is consistent with the high accuracy of the most-frequent class baseline for POS tagging.

The trends in the capacity and geometry were similar across different Transformer architectures. The manifold capacities increase across the downstream layers, mediated by the reduction in the manifold dimension, radius, and center correlations in masked data, (Fig.~\ref{fig:mft_manymodels}), as demonstrated by the POS manifolds. 
In unmasked data, the capacity is higher overall compared to the capacity in masked data, and the manifold dimension, radius, and correlations are overall smaller, due to the existence of the strong signal in the input. The capacity of unmasked manifolds decrease, and this rate of decrease is steep in Word manifolds, but gradual in linguistic manifolds  due to the contextualization effect (see SM~\ref{SI:ling:tasks:models} for  the rest of the linguistic manifolds across models). 

\begin{figure}[h]
\begin{center}
\centerline{\includegraphics[width=\columnwidth]{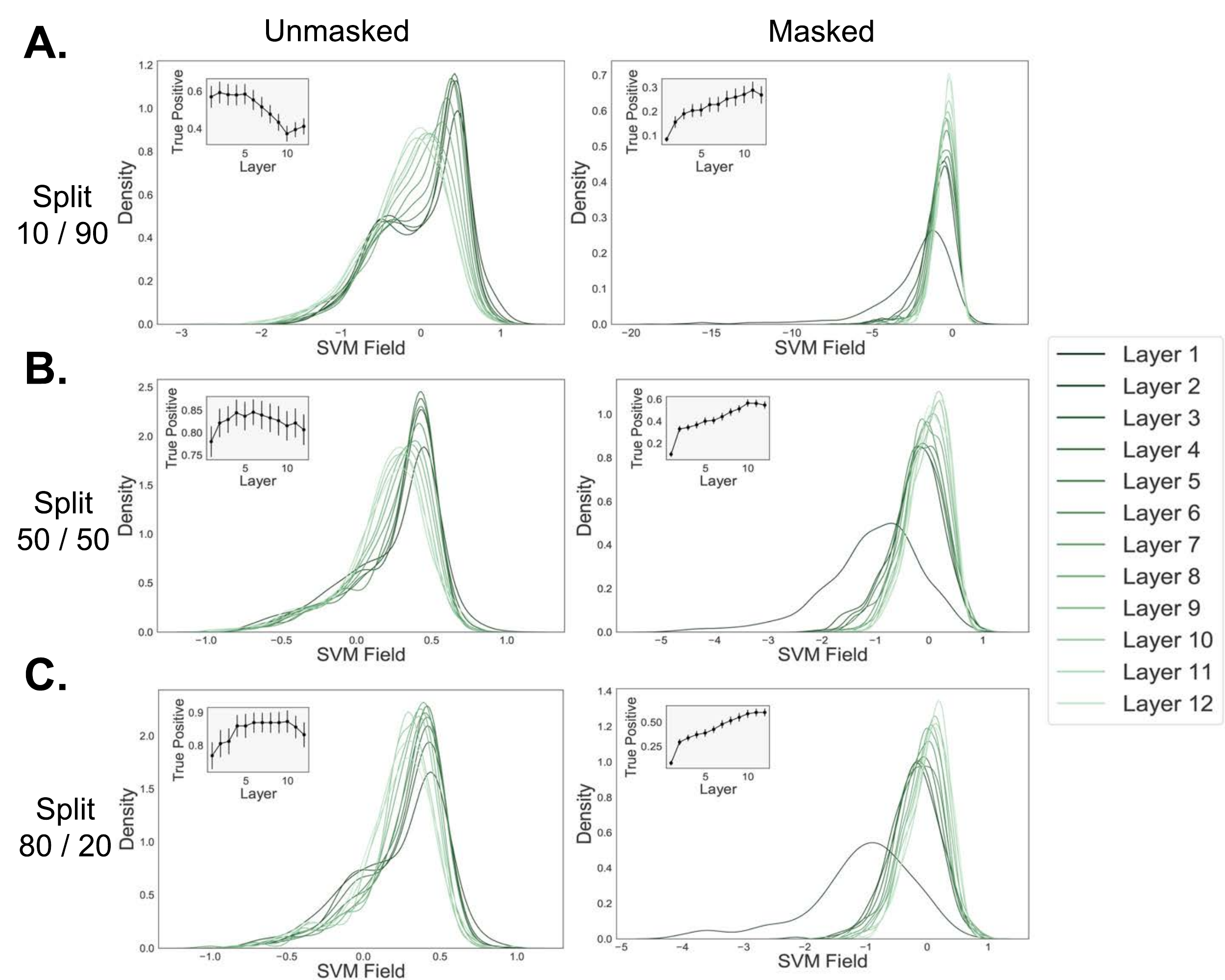}}
\vspace{-2mm}
\caption{\textbf{SVM Field Distribution reflects Emergence of linearly separable language manifolds} Fields distribution of POS Manifolds on (top left) unmasked  and (top right) masked data, on center-normalized data with (Top) 80\% Tr/ 20\% Test Split, and (Bottom) 10\% Tr/ 90\% Test Split, on conditioned datasets (true positive + false negative). (Insets in each figure shows true positive rate). 80/20 case is consistent with \cite{liu_et_al}, and 10/90 on unmasked data shows a different trend.} \vspace{-8mm}
\label{fig:SVMfields}
\end{center}
\end{figure}

\vspace{-2mm}
\paragraph{SVM fields statistics analysis}

We supplement the manifold capacity analysis by analyzing the statistics of the fields, i.e., the signed perpendicular distances between feature vectors and the SVM hyperplane, as described in Sec.~\ref{sssection:method_SVMfields}. Fig.~\ref{fig:SVMfields} shows this for POS manifolds for different train-test splits, and across the BERT model layers. In the masked data (Fig. \ref{fig:SVMfields}, right column), across all train/test splits, the peak of the fields distribution moves positively away from the origin from early to downstream layers, while the width gets smaller, meaning an increase in the signal to noise ratio, which is also reflected in the measured accuracy (Fig.~\ref{fig:SVMfields}, masked, Inset Figures). 
In the unmasked data, the peak of the fields distribution moves from the positive side to the origin (in the negative direction), across different train/test split regimes, showing the decreased separability, consistent with the trends observed in the manifold capacity. Interestingly, we find that the trend in linear separability, as measured by the accuracy across layer depth, is dependent on the size of the training set. With a train/test split of 10/90, the fraction of positive fields (i.e. accuracy) decreases across layers (Fig. \ref{fig:SVMfields}, Top Left Inset). On the other hand, when we use the same train/test split of 80/20 used by \citet{liu_et_al}, we recover their observation that the fraction of positive fields increases across the layers.

\begin{figure*}[h]
\begin{center}
\centerline{\includegraphics[width=1.75\columnwidth]{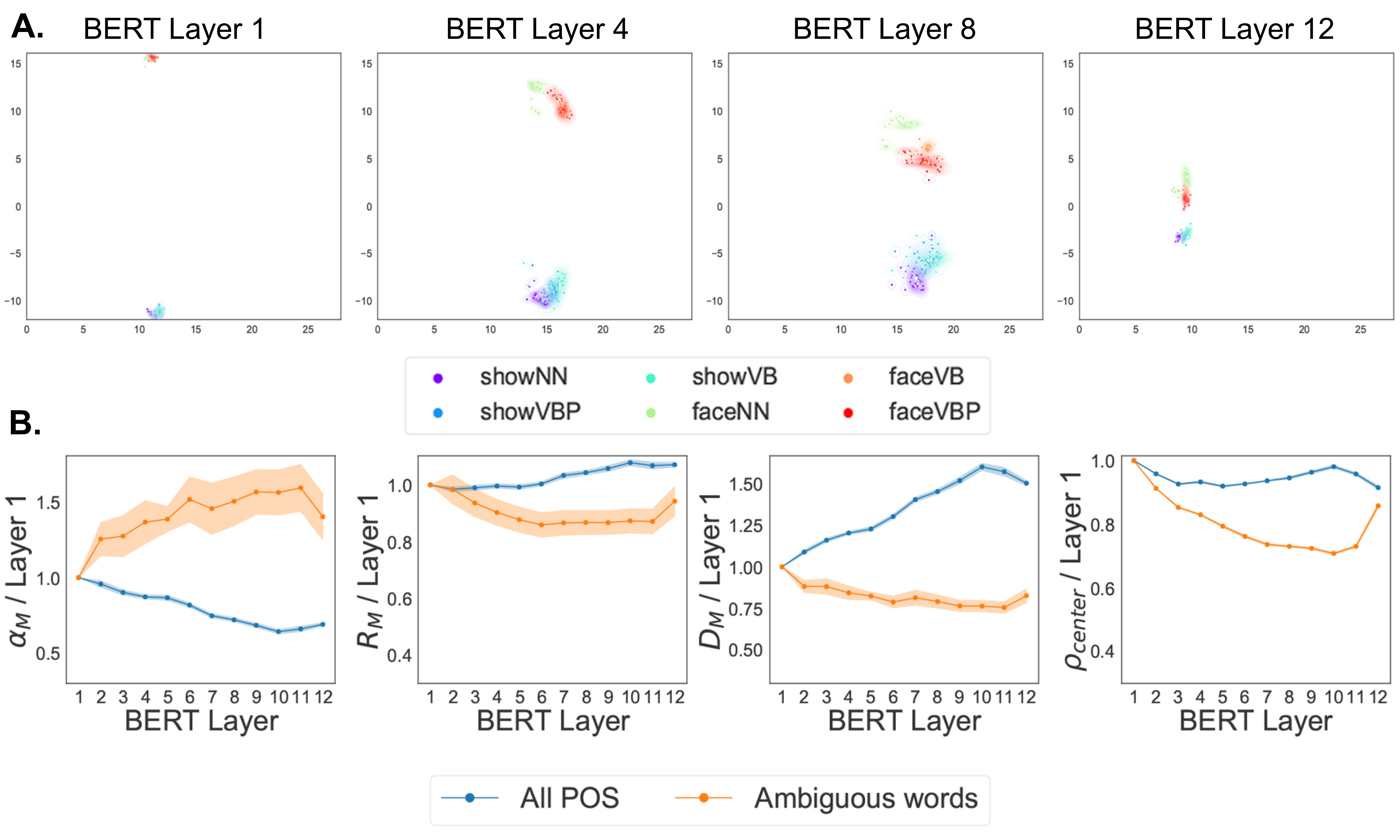}}
\caption{\textbf{Visualization and geometric property of POS manifolds with ambiguous words dataset}, unmasked data. (A) PCA visualization of ambiguous words that have multiple possible POS tags across BERT layers. (B) Normalized manifold capacity, radius, dimension, and center correlations of POS manifolds defined on ambiguous words versus on random sampling. For ambiguous words, shaded error bar is calculated from the mean of manifold metrics (including 5 random seeds from the ambiguous word dataset). 
}
\label{fig:PCA_words_ambiguous}
\vspace{-7mm}
\end{center}
\end{figure*}

\subsection{Geometric Evidence for the Stronger Effect of Context in Ambiguous Words}
\label{ssec:visualization}

Following our analysis of word and linguistic manifolds with metrics from both manifold theory and SVM fields, we searched for evidence of co-evolution of separable language manifolds of multiple types by visualizing their representations using PCA. We focus our analysis on two specific data: (1) words that occur across multiple POS tags, which we call "ambiguous words", and (2) open vs. closed POS tags \cite{lyons1977semantics}, corresponding to POS tags that have fixed class membership (closed tags) and usually includes only a few words versus those that accept new members (open tags) and usually contains many words.

\paragraph{Geometry of ambiguous words} Much of the POS information can be inferred by the choice of a word, regardless of the context, as some words always correspond to a specific POS (e.g., "the", "a", etc.). To test the additional information about POS tags gained by the neighboring context, we focus our analysis here on a specific set of words which occur across at least three different POS tags. This analysis is particularly useful because even for word manifolds, there is an effect of contextualization, observed by the steep decrease in the capacity across layers. Can we reduce the effect of underlying word manifolds by specifically choosing ambiguous words with multiple POS tags? The hypothesis here is that the amount of untangling of POS information might depend on how ambiguous underlying words are. 

To test this, we first visualize word embeddings with multiple POS tags with PCA. We find that in the first layer, while the words are highly separated, different POS vectors are embedded very close to each other, and as the representations travel downstream, the POS sub-clusters within a same word (different colors) separate, while the overall manifold sizes get larger, clearly showing the competing effect of contextualization. This is a known challenge in the "untangling" deep networks, observed in vision; that is, the network needs to embed the whole data in high dimensional space, while separating and compressing the categorical data in low dimensional space, all using the same network parameters \cite{cohen2019separability, recanatesi2019dimensionality}.  

\begin{figure}[h]
\begin{center}
\centerline{\includegraphics[width=\columnwidth]{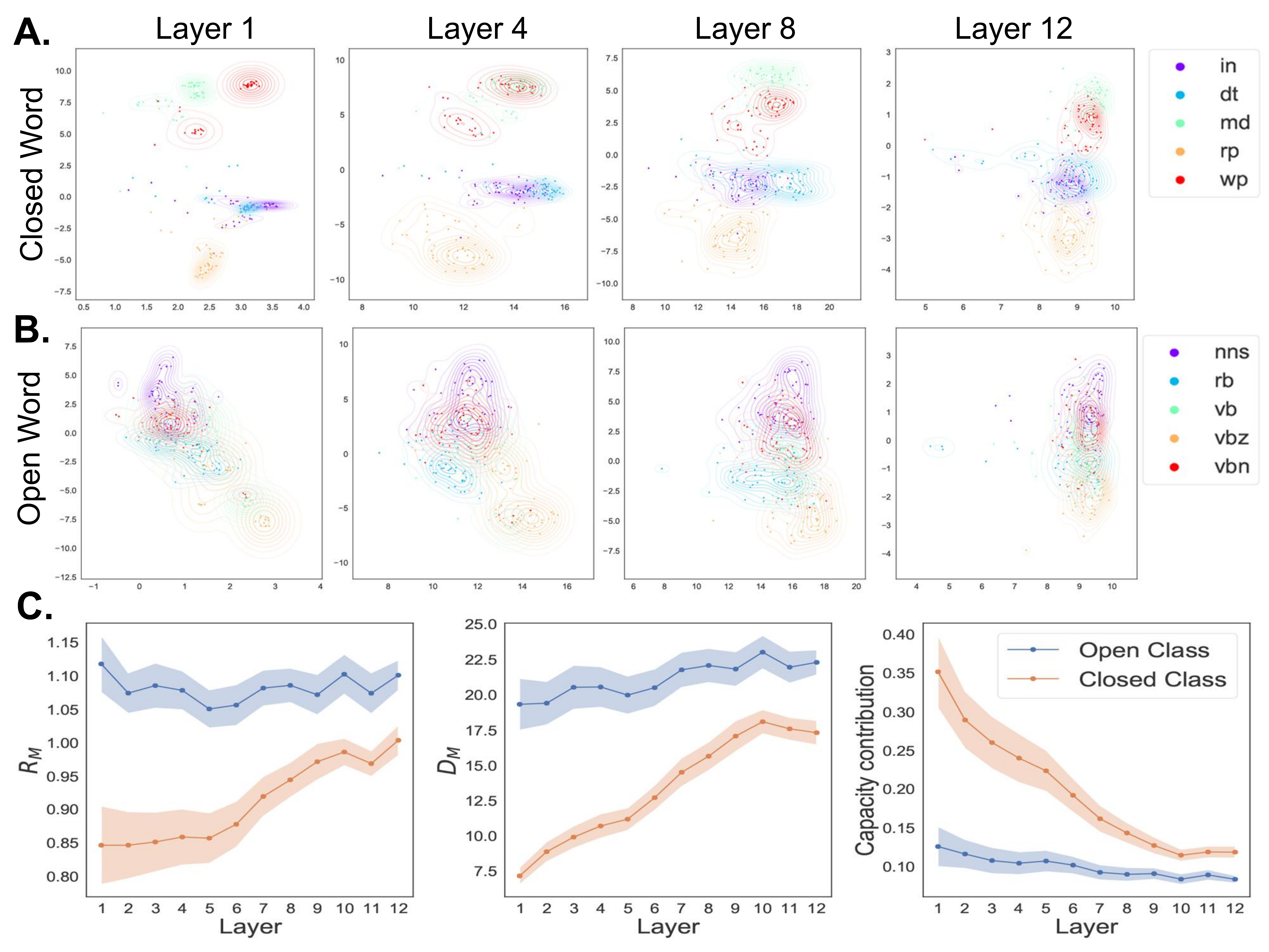}}
\caption{\textbf{Linguistic manifold visualization and geometric property of closed and open word class} PCA visualization for 5 randomly selected POS classes from open word class (A) and closed word class (B) (density is estimated by bivariate kernel density estimate). (C) Manifold radius, dimension and capacity contribution of POS classes in open word class and closed word class across BERT layer. Error bar represents standard error between different POS classes within open word or closed word class. 
}
\label{fig:PCA_POS_closed_open}
\vspace{-10mm}
\end{center}
\end{figure}

With the competing effect of contextualization on manifold geometry observed by PCA visualization, we then quantify these effects with our manifold analysis metric. One question is: while it's clear that the POS information is segregating conditioned on each word (Fig. \ref{fig:PCA_words_ambiguous}, Top), are POS manifolds defined across many words also segregating, and do they segregate more, if the underlying words are ambiguous? Our results suggest that indeed BERT layers untangle POS manifolds from ambiguous words more than they untangle typical POS manifolds. In Fig. \ref{fig:PCA_words_ambiguous}B, while POS manifold capacity of randomly sampled dataset decreases (blue line), POS manifold capacity of manifold sampled from ambiguous words increase across layers (orange line). Furthermore, we also observe that in the ambiguous dataset, manifold radius, dimension and center correlation also decrease across layers, suggesting a trend of untangling \cite{cohen2019separability, stephenson2019untangling}, as opposed to the entangling trend in randomly sampled POS dataset. Interestingly, the untangling effect is strongest in the middle layers, which is consistent with the PCA visualization (Fig.\ref{fig:PCA_words_ambiguous}). This effect maybe the result of the model objective to predict the word identity in the last layer, making several POS manifolds within a word becomes more entangled. 


We observe that this analysis can explain the ostensible contradiction between the decrease in manifold capacity across layer depth that we measured and the increase in accuracy across layer depth that has generally been observed from supervised probes \cite{liu_et_al}. In particular, the visualizations indicate that there is an \emph{overall} entangling of representations in later layers (when averaged across all words), leading to decrease in manifold capacity. However, there is an untangling of representations \emph{within} each word, contributing to higher probe accuracy in later layers.

\paragraph{Geometry of open vs. closed POS tags}

\begin{figure}[h]
\begin{center}
\centerline{\includegraphics[width=\columnwidth]{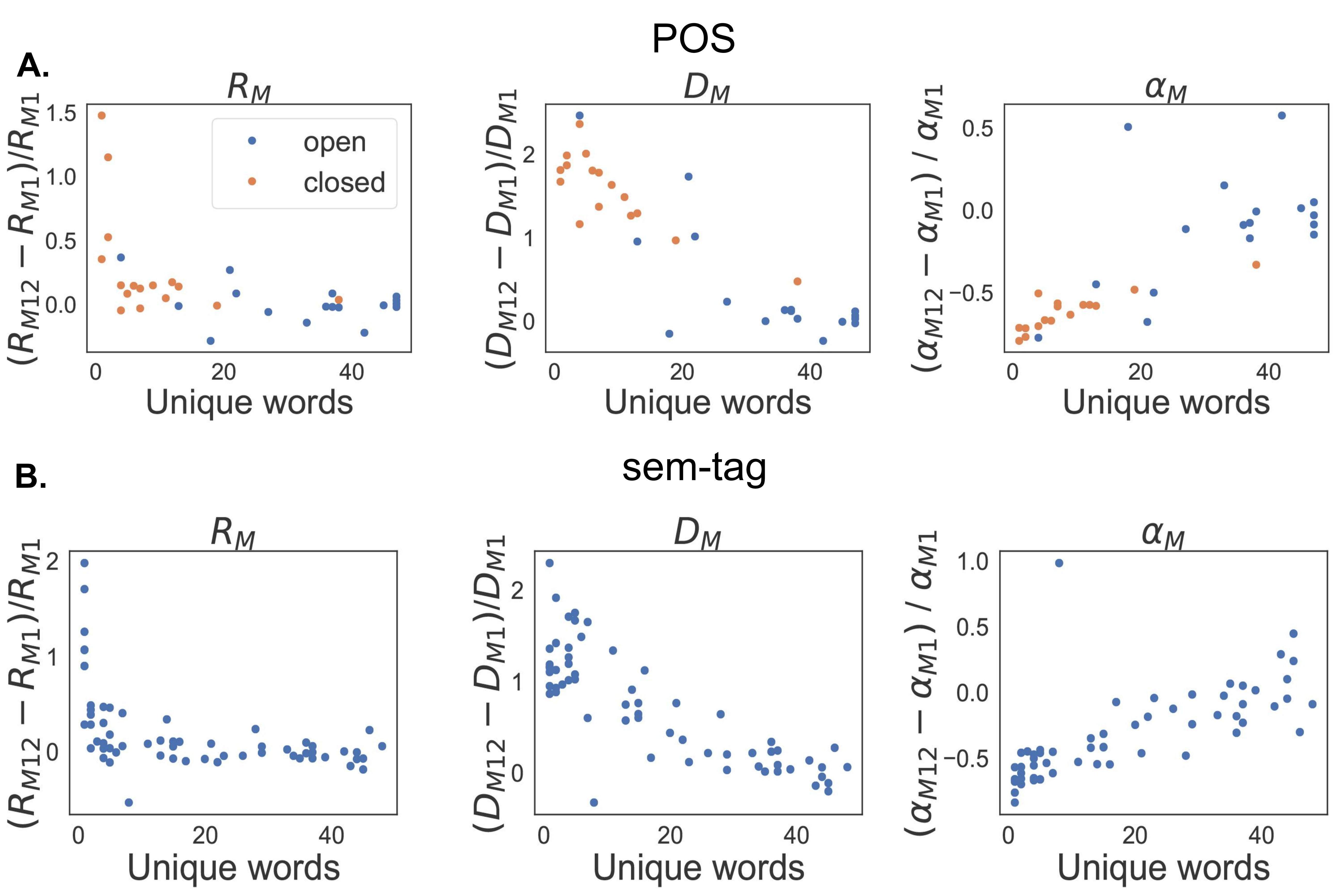}}
\caption{\textbf{Relationship between manifold geometrical evolution and number of unique words within each manifold} Relationship between manifold radius, dimension and capacity ratio to the number of unique word in each manifold. The ratio is calculated by the formula $(X_{12}-X_1)/X_1$ with $X_1$ and $X_{12}$ be the measurement from layer 1 (first layer) and layer 12 (last layer) respectively.}
\label{fig:POS_unique words}
\vspace{-10mm}
\end{center}
\end{figure}

In addition to ambiguous words, we analyze POS classes with different number of words within each POS class, in unmasked data. We compare the geometry of closed class POS categories, corresponding to a small number of distinct words, and open class POS categories, with a large number of distinct words. Do closed-class POS manifolds show a similar geometrical transformation properties as word manifolds? Open-class manifolds, as embedded across many words, might be highly entangled in the input layer; do they show more emergent separability compared to closed-class manifolds? 

Based on their PCA visualizations (Fig. \ref{fig:PCA_POS_closed_open}A-B)), closed word POS classes are indeed already well separated in the embedding layer, and over the layers, these POS manifolds become more tangled, similar to the word manifolds trends seen previously (Fig. \ref{fig:capacity_bert_ntasks}B). On the other hand, open POS classes are quite entangled in the beginning embedding layer, and their change in separability appears to be relatively small (Fig. \ref{fig:PCA_POS_closed_open}B). Applying our manifold capacity method clearly shows that the measured manifold dimension and radii increases across layers in the closed class, but changes minimally for the open class (Fig. \ref{fig:PCA_POS_closed_open}C). Similarly, the capacity contribution \footnote{Manifold capacity is defined as $\alpha = \langle \alpha_\mu^{-1} \rangle_\mu^{-1}$, where $\alpha_\mu$ is a capacity contribution from the $\mu$ th manifold. For capacity contribution for open (closed) classes (Fig. \ref{fig:PCA_POS_closed_open}C), $\mu$ refers to manifold indices corresponding to open (closed) classes.} decreases significantly for closed class, but changes minimally for the open class. 

Finally, we study the relationship between manifold geometry of output/input ratio (defined in Fig. \ref{fig:POS_unique words} caption) and number of unique words within a linguistic manifold for POS and sem-tag tasks (Fig. \ref{fig:POS_unique words}). While manifold capacity ratio is correlated with the number of unique words, manifold radius and dimension ratio is anti-correlated. This result implies that for POS and sem-tag, the linguistic manifolds with large number of words counteract the effect of expanding underlying word manifolds the most, signifying the structural evidence of contextualization on untangling of POS and sem-tag representations.


\subsection{Geometry of Learning Dynamics and Task-transferability}

\begin{figure}[h]
\begin{center}
\centerline{\includegraphics[width=\columnwidth]{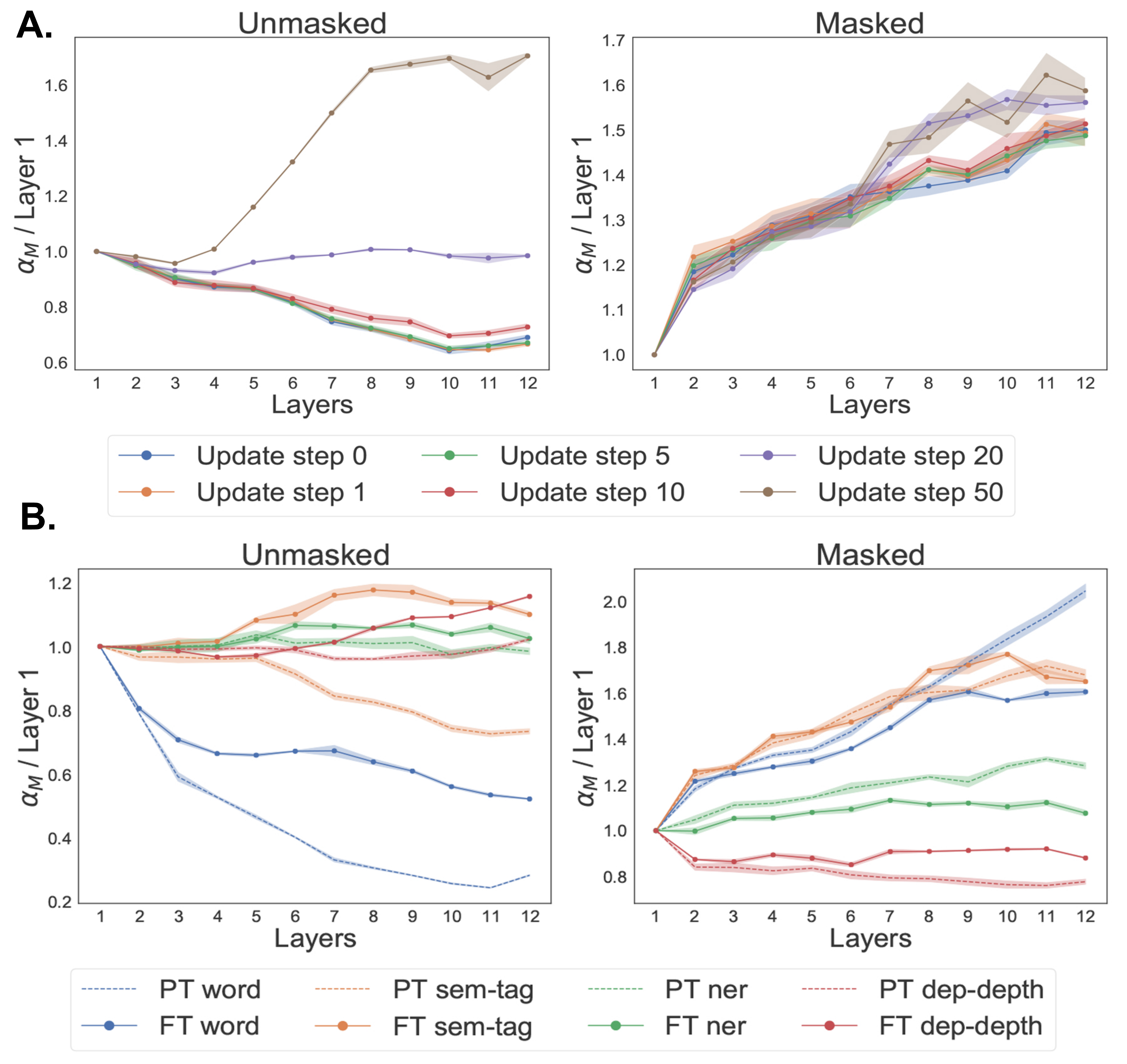}}
\caption{\textbf{Geometric evidence of fine tuning and task transfer representations} (A) Emergence of POS manifolds on (left) unmasked, (right) masked data across update steps of POS fine-tuning on BERT. (B) Linguistic manifolds on (left) unmasked, (right) masked data on (solid) final step of POS fine-tuning on BERT, compared with pretrained BERT (dashed).}
\label{fig:mft_transfer}
\vspace{-3mm}
\end{center}
\end{figure}

In addition to evaluating the MFTMA metrics on the pre-trained networks, the analysis can also be done as the training progresses. Here, we trained a popular fine-tuning task, POS, on a pre-trained BERT, and trace the geometry of learned representations over the course of the training. Fig.~\ref{fig:mft_transfer}A shows the evolution of representation geometry in different stages of learning epochs, measured by the capacity of manifolds measured on the same tags as the training task (POS). Across training steps, POS manifolds in contexualizing regime (unmasked data) become more separable across layers, resulting in the regime where the manifold capacity increases across layers in the final step. Surprisingly, POS fine-tuning has little effect on masked data.

Furthermore, we characterize how these representations fine-tuned with POS task transfer to a different task, by measuring the manifold capacity with other linguistic tags, i.e., word, ner, sem-tag, and dep-depth. In unmasked data, the linguistic manifolds increase in their overall capacities with POS fine-tuning, showing the evidence of the task transfer. In masked data, most linguistic manifolds also improve in their capacity by a small amount, with an exception of word and NER manifolds, where an overall entanglement is observed between pretrained and fine-tuned models. Note that masked token is never seen during fine-tuning.

In addition, we supply the analysis on the manifold capacity vs. task performance (F1, precision, recall) across the fine-tuning updates in SM~\ref{SI:fine:tune}.

\section{Discussion}

In this paper, we studied the emergent geometric properties of word and linguistic object manifolds and their linear separability, as measured by the shattering manifold capacity. Across different network models and datasets, we find that language manifolds emerge across the layers of the hierarchy. Particularly in the predictive manifolds defined on the masked data, the manifold capacity consistently increases, similar to the 'untangling' phenomena observed in visual and auditory sensory systems~\cite{cohen2019separability, chung2020separable,stephenson2019untangling}. Contextualized  manifolds defined on unmasked words also show implicit emergence of separable linguistic geometry, counteracting the strong entangling effect of word manifolds. Interestingly, the untangling effect of contextualization is stronger in words and linguistic tags that are ambiguous. 

The emergence of increasing manifold capacity (accompanied by the emergent manifold geometry) on the masked data is reflective of the fact that the representations emerge across layers to become more "similar" to the last layer's output, consistent with the prior reports on the increasing mutual information between the intermediate and the last layers of contextual embedding models \cite{voita2019}. The decrease of the word manifold capacity on unmasked data implies that information about the input word generally gets lost as representations get transformed downstream, similar to the reduced mutual information between input and intermediate layers \cite{voita2019}. Furthermore, our metric goes beyond the constraints of the comparative measure, as the manifold capacity measures the amount of object information for any categories, allowing analysis of higher level linguistic categories such as POS. 

Contextualization is a competition between gaining information from neighboring words, without losing information about the original word. Just as much as the original word representation is enhanced by the context, the same word representation is used to enhance the representation of other words. Measuring how distributed systems such as Transformers balance this multitude of information flows and implement linguistic information in a mixed representation is a theoretical challenge. Enabled by the recent manifold analysis technique, we report the quantifiable structural evidence of the evolution of language manifolds in connection to their linear separability in widely used language models. 

Our methodology and results suggest many interesting future directions. We hope that our work will motivate: (1) the theory-driven geometric analysis of emergent representations underlying complex tasks such as prediction and contextualization; (2) the development of new theoretical frameworks that link the representation geometry and tasks with underlying causal structures; (3) the future study of language representations in the brain via the lens of geometry.

\pagebreak
\section*{Acknowledgements}
We thank Haim Sompolinsky, Larry Abbott, Ev Fedorenko, Roger Levy, Greg Wayne, Jon Gauthier and Abhinav Ganesh for helpful discussions. The work was funded by Intel Research Grant, NSF NeuroNex Award DBI-1707398, and The Gatsby Charitable Foundation.

\bibliography{ms}
\bibliographystyle{icml2020}

\clearpage
\setcounter{section}{0}
\setcounter{figure}{0}
\section{Supplementary Material}
\subsection{Empirical manifold capacity and theoretical manifold capacity}
\subsubsection{Empirical manifold capacity}
In this section, we provide detailed description about how to find empirical manifold capacity. Given $P$ object manifolds, $N_c$, the critical number of feature dimensions, is defined as the necessary number of feature dimensions so that $P$ object manifolds, with randomly assigned $+/-$ labels for each manifold, can be linearly separated half the time on average (see \cite{stephenson2019untangling}). The empirical manifold capacity is defined as $P/N_c$, which is the ratio between number of object manifold and the critical number of feature dimensions. To find $N_c$, a bisection search is performed until either the linearly separated fraction is within an error tolerance range $\epsilon = 0.05$ or the number of iteration exceeds $100$. If the number of feature dimensions $N$ is larger than $N_c$, then the fraction of linearly separable dichotomies is close to $1$, and the data is in the linearly separable regime. Conversely, if the number of feature dimensions $N$ is smaller than $N_c$, then the fraction of linearly separable dichotomies is close to $0$, and the data is in the linearly inseparable regime. 

In our experiments, we first randomly sample 20 instances for each manifold to perform the analysis. Then, for each candidate feature dimension in the bisection search, we sample $51$ randomly assigned dichotomies to compute the linearly separable fraction. We use features extracted from pre-trained {\tt bert-base-cased} model. Note that we exclude the embedding layers in this analysis due to the overlapping data point between manifolds as reported in Section~\ref{sec:datasets}.

\subsubsection{Theoretical manifold capacity}
Theoretical capacity used here is Mean-Field Theoretical Manifold Capacity described in Section~2 of the main text. We use $\kappa = 10^{-8}$ and $n_t = 300$, in which $\kappa$ is the margin size and $n_t$ is the number of Gaussian vectors to sample per manifold (see \cite{chung2018classification}). We also use the same randomly chosen 20 instances from the simulation capacity analysis for each manifold. 

\begin{figure}
\centerline{\includegraphics[width=\columnwidth]{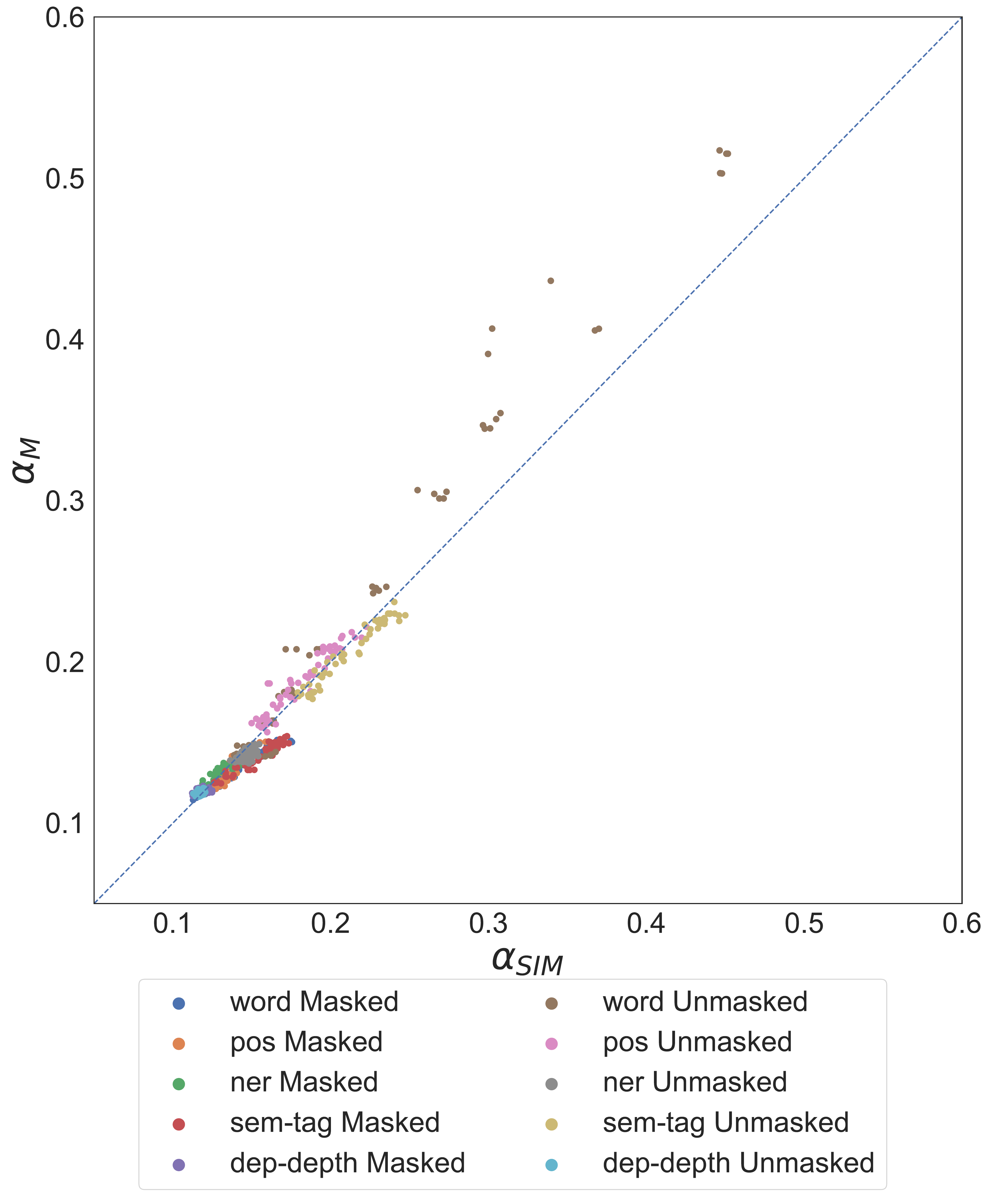}}
\caption{Simulation capacity vs. MFT capacity in {\tt bert-base-cased} model.} 
\label{fig:sim_cap}
\end{figure}

Figure~\ref{fig:sim_cap} shows a close match between simulation capacity and the MFT manifold capacity observed in various linguistic tasks, measured across the hierarchy of {\tt bert-base-cased} model. 

\subsection{Model architecture details}
\subsubsection{Pre-trained Models Details}
We present briefly the pre-trained models that we used for the experiments.
\begin{itemize}
    \item {\bf BERT} {\tt bert-base-cased}. 12-layer, 768-hidden, 12-heads, 110M parameters.
    \item {\bf RoBERTa} {\tt roberta-base}. 12-layer, 768-hidden, 12-heads, 125M parameters.
    \item {\bf ALBERT} {\tt albert-base-v1}. 12 repeating layers, 128 embedding, 768-hidden, 12-heads, 11M parameters.
    \item {\bf DistilBERT} {\tt distilbert-uncased}. 6-layer, 768-hidden, 12-heads, 66M parameters.
The model distilled from the BERT model {\tt bert-base-uncased} checkpoint.
    \item {\bf OpenAI-GPT} {\tt openai-gpt}. 12-layer, 768-hidden, 12-heads, 110M parameters.
\end{itemize}
For each pre-trained model, input text is tokenized using its default tokenizer and features are extracted at token level.

\subsubsection{Fine-tuned Model Details}
We fine-tuned BERT {\tt bert-base-cased} model on POS downstream task with the following hyper-parameters:
\begin{itemize}
    \item Epsilon for Adam optimizer: $1\mathrm{e}{-8}$.
    \item Initial learning rate for Adam: $5\mathrm{e}{-5}$.
    \item Max gradient norm: 1.
    \item Maximum total input sequence length after tokenization: 128. Longer sequences are truncated and shorter sequences are padded.
\end{itemize}

\subsection{Datasets and Manifolds Details}
\label{sec:datasets}
In this section, we provide some information about the labels defining the manifolds for each task with some additional details (e.g., overlapping).

\subsubsection{Word}
Labels are the following: {\tt the, of, to, in, and, for, that, is, it, said, on, at, by, as, from, with, million, was, be, are, its, he, but, has, an, will, have, new, or, company, they, this, year, which, would, about, says, market, more, were, his, billion, had, their, up, one, than, some, who, been, stock, also, other, share, not, we, when, last, if, years, shares, all, president, first, two, sales, after, inc., because, could, out, trading, there, only, business, do, such, can, most, into}.

Note that, by definition, there is no overlapping between the manifolds.

\subsubsection{Part-of-Speech}

Labels are the following: {\tt NN, IN, NNP, DT, JJ, NNS, CD, RB, VBD, VB, CC, TO, VBZ, VBN, PRP, VBG, VBP, MD, POS, PRP\$, WDT, JJR, NNPS, RP, WP, WRB, JJS, RBR, EX, RBS, PDT, FW, WP\$}.

Labels are described in \url{https://www.ling.upenn.edu/courses/Fall_2003/ling001/penn_treebank_pos.html}.

There is $0.032\%$ of overlapping pairs of words in the embedding layer due to the occurrence of a same word at the same position in multiple sentences with a different POS label. However, as expected, there is no overlapping for higher layers. \\
\\
For the POS open-word class and closed-word class analysis, we used the following assignment of POS tags:
\begin{itemize}
    \item Open-word class: {\tt JJ, JJR, JJS, RB, RBR, RBS, NN, NNS, NNP, NNPS, VB, VBD, VBG, VBN, VBP, VBZ, FW}
    \item Closed-word class: {\tt IN, DT, CD, CC, TO, PRP, MD, POS, PRP\$, WDT, RP, WP, WRB, EX, PDT, FW, WP\$}
\end{itemize}

For the ambiguous words analysis, we used the following words with associated POS tags:
{\tt back (RP, RB, JJ, NN), cut (VBN, VBD, NN, VB), set (VBD, VB, NN, VBN), close (NN, RB, JJ, VB), lower (RBR, VB, JJR),
closed (VBD, VBN, JJ), estimated (JJ, VBD, VBN), call (NN, VB, VBP), come (VB, VBN, VBP), earlier (JJR, RBR, RB), pay (VB, VBP, NN), up (RP, RB, IN), over (IN, RB, RP), proposed (JJ, VBN, VBD), face (VBP, VB, NN), continued (JJ, VBD, VBN), down (IN, RB, RP), show (VB, VBP, NN), off (RP, RB, IN),
better (JJR, RBR, RB), longer (RBR, RB, JJR), half (NN, PDT, DT), expected (VBN, JJ, VBD),
buy (VB, NN, VBP), look (VB, NN, VBP)}

\subsubsection{Semantic Tags}

Labels are the following: {\tt CON, REL, IST, DEF, LOC, PST, ORG, PER, DIS, SUB, EXS, NOW, PRO, HAS, AND, EXG, EXV, QUA, GPE, EXT, ENT, TIM, COO, APP, EPS, YOC, FUT, DOM, NOT, MOR, MOY, ENG, INT, TOP, ALT, ENS, ETV, POS, PRX, BUT, EPT, UOM, DST, QUE, NEC, EPG, IMP, ART, HAP, ETG, ROL, DOW, SCO, REF, COM, DEC, EXC, NAT, RLI, LES, EFS}.

Labels are described by \citet{abzianidze2017towards}.

Note that there is no overlapping between the manifolds.

\subsubsection{Named-Entity Recognition}
The NER dataset includes 18 labels described by \citet{weischedel2011ontonotes}, consisting of 11 types ({\tt GPE, LOCATION, WORK\_OF\_ART, EVENT, LAW, PRODUCT, LANGUAGE, PERSON, ORG,  NORP, FAC}) and 7 values ({\tt DATE, PERCENT, CARDINAL, TIME, QUANTITY, ORDINAL, MONEY}). With BIO tagging scheme, each label can occur either with {\tt B-} ({\it beginning}) prefix or with {\tt I-} ({\it inside}) prefix; there is an additional {\tt O} ({\it outside}) label for words that are not named-entities. 

There is $0.014\%$ of overlapping pairs of words in the embedding layer due to the occurrence of a same word at the same position in multiple sentences with a different NER label. However, as expected, there is no overlapping for higher layers.

\subsubsection{Dependency Depth}

We select dependency depths from 0 to 21. From depth 18 to 21, we have respectively 12, 12, 5, 4 samples occurring in the corpus (instead of 50 for other depths).

Note that there is no overlapping between the manifolds.

\begin{figure*}[ht]
\begin{center}
\centerline{\includegraphics[width=2\columnwidth]{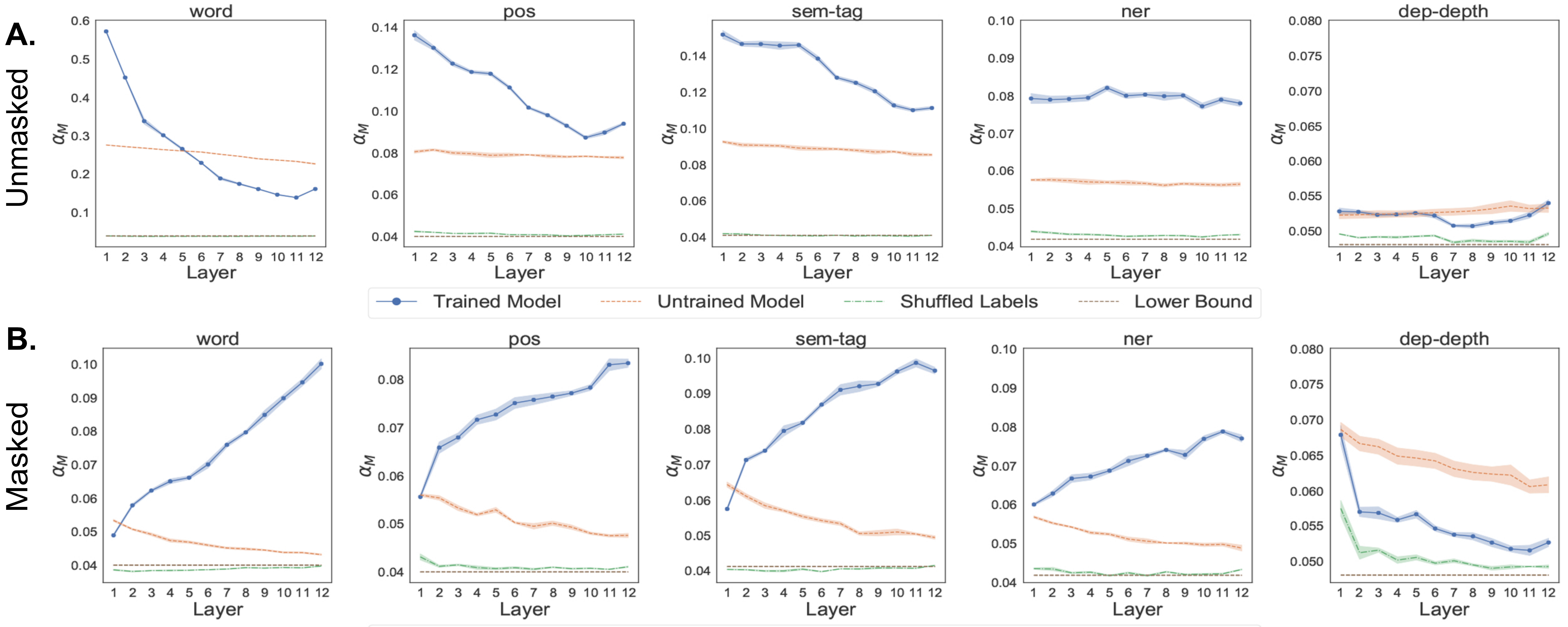}}
\caption{Randomly controls for manifold capacity in {\tt bert-base-cased} model.} 
\label{fig:rand_init}
\end{center}
\end{figure*}

\subsection{Additional Experiments}
\subsubsection{Random baseline control for manifold capacity}
\label{SI:random}
We compare in Figure~\ref{fig:rand_init} the manifold capacity to three different manifold capacity baselines: 
\begin{itemize}
\item {\bf Lower bound}. The lower bound capacity $LB$ is defined as the classification capacity of unstructured manifolds and only depends on the number of samples in each manifold. 
\begin{equation}
LB = \frac{1}{\frac{1}{n}\sum^{n}_{i=1}\frac{l_i}{2}}
\end{equation}
where $LB$ is lower bound capacity, $n$ is the number of manifolds, $l_i$ is the number of samples in manifold $i$ (see \cite{stephenson2019untangling}).
\item {\bf Randomly initialized (untrained) model}. All model weights are set to a random number. Note that this random initialization has also an impact on the embedding layer. 
\item {\bf Shuffled label manifolds}. The manifolds are shuffled without repetition and the number of samples for each manifold are preserved.
\end{itemize}

For both masked and unmasked data from {\tt bert-base-cased} model, the capacity of shuffled label manifolds matches closely with the lower bound capacity, suggesting that randomly assigned manifold in different layers and linguistic tasks follow closely with the lower bound capacity. 

Concerning the untrained model with random weights, in unmasked data, the capacities in the embedding layer are higher than lower bound and lower than the capacities in the pre-trained model. This reflects the fact that word vectors are already somewhat separated in the embedding layer, and the random weights don't improve or decrease the capacity. 
For the masked data with untrained model, the manifold capacity decreases across layers. The trends observed here are similar to prior work by \citet{jawajar2019structure}. Note that as observed by \citet{gaier2019weight}, structured manifolds could emerge even in untrained models.

\subsubsection{Analysis of Raw SVM Fields Distribution of POS manifold}
\label{SI:SVM}

\begin{figure*}
\begin{center}
\centerline{\includegraphics[width=2\columnwidth]{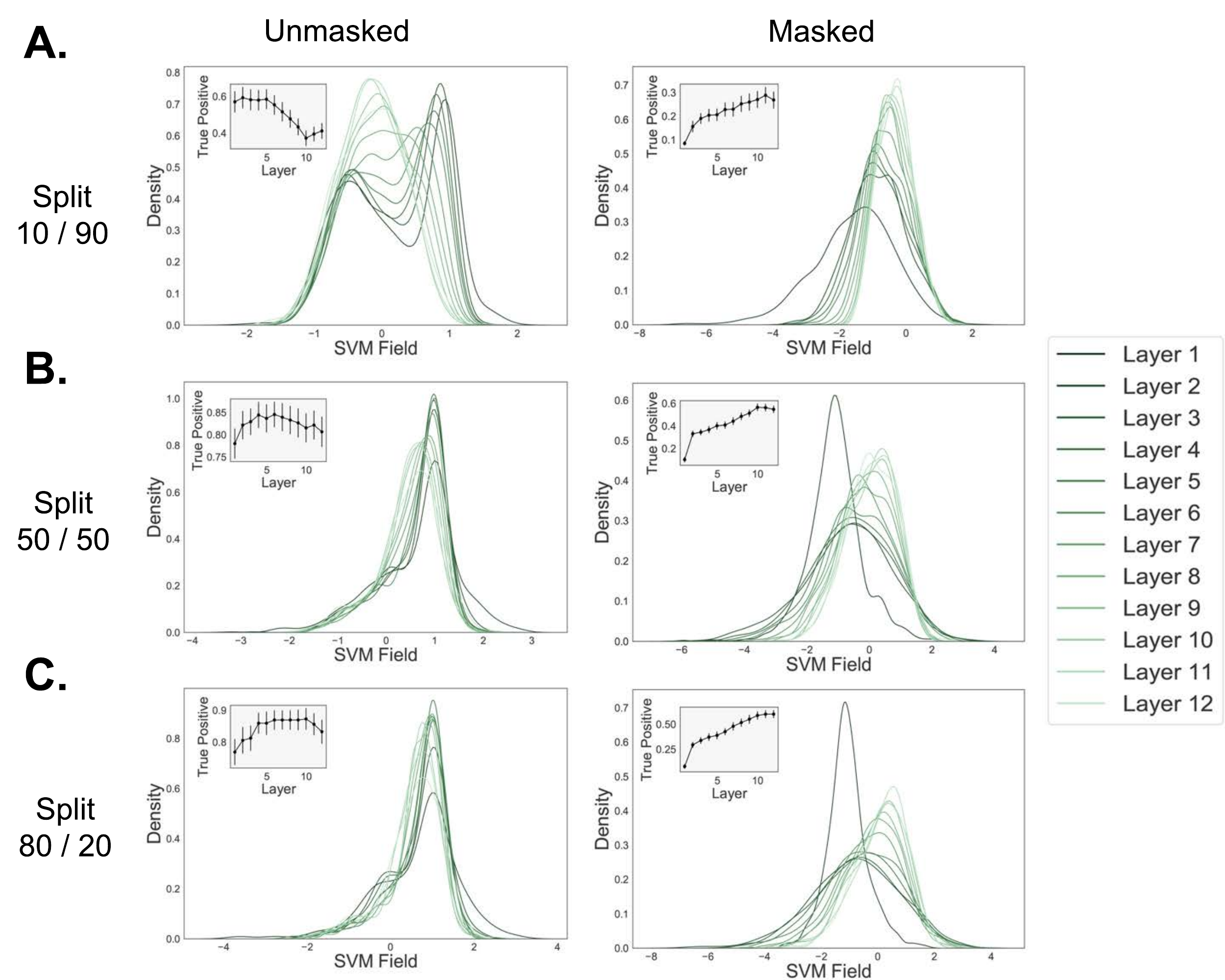}}
\caption{Raw SVM fields of POS manifold with {\tt bert-base-cased} model.} 
\label{fig:svm_field}
\end{center}
\end{figure*}

We report in Figure~\ref{fig:svm_field} the raw SVM fields distribution of POS manifold with {\tt bert-base-cased} model.
The raw SVM fields distribution, despite of having a different distribution shape, shows similar trend across layers with the normalized SVM fields distribution described in the main text for both masked and unmasked dataset. The accuracy for raw SVM field distribution matches exactly the accuracy for normalized SVM fields distribution because normalization doesn't change sign of the fields. For unmasked data, the peak of the field distribution and the right tail moves slightly to the negative direction in all different train/test splits. For masked data, although the peak shifts to the negative direction, the right tail of the distribution extends to the positive direction in all different train/test splits, representing an increase in accuracy across layers. 

\begin{figure*}[ht]
\begin{center}
\centerline{\includegraphics[width=2\columnwidth]{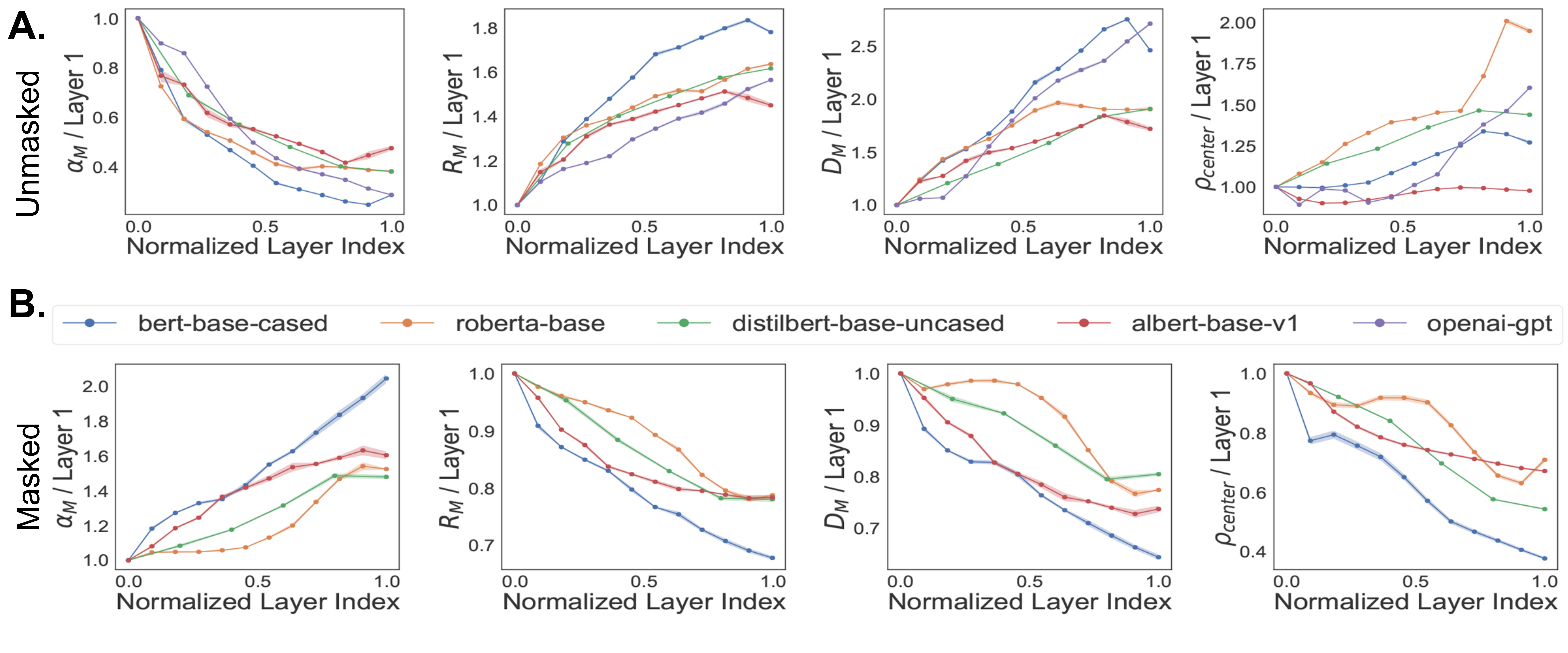}}
\caption{Geometric properties of word manifold in different models.} 
\label{fig:fig4_word}
\end{center}
\end{figure*}
\subsubsection{Geometric Properties Evolution through Sequential Layers across Linguistic Tasks and Models (Additional Figures)}
\label{SI:ling:tasks:models}
We report geometric properties (manifold capacity, radius, dimension and center correlation) for word, semantic tags, NER and dependency depth manifolds for the different models.

\paragraph{Word}

For word manifolds, as reported in Figure~\ref{fig:fig4_word}, similarly to POS manifold, the capacity increases for unmasked data and decreases for masked data in all the different models. In both masked and unmasked cases, the trend is clear and steep. In the masked case, the inputs are masked and feature vectors values only depend on the positional embedding and are not related to the word strings; since the model is trained to predict the masked word token, the word manifolds emerge across layers. In the unmasked case, the inputs are context-free embedding word vectors and are well separated; since the model tries to contextualize the word using its neighbor words, the word manifolds get entangled and lead to a decrease in word manifold capacity. The radius, dimension and center correlation measures also reflect the observed trend in the capacity. In the unmasked data, the radius, the dimension and center correlation of word manifolds increase across layers, representing manifold entangling. In the masked data, the dimension, radius and center correlation decrease across layers, suggesting manifold untangling.

\paragraph{Semantic Tags}
\begin{figure*}[ht]
\begin{center}
\centerline{\includegraphics[width=2\columnwidth]{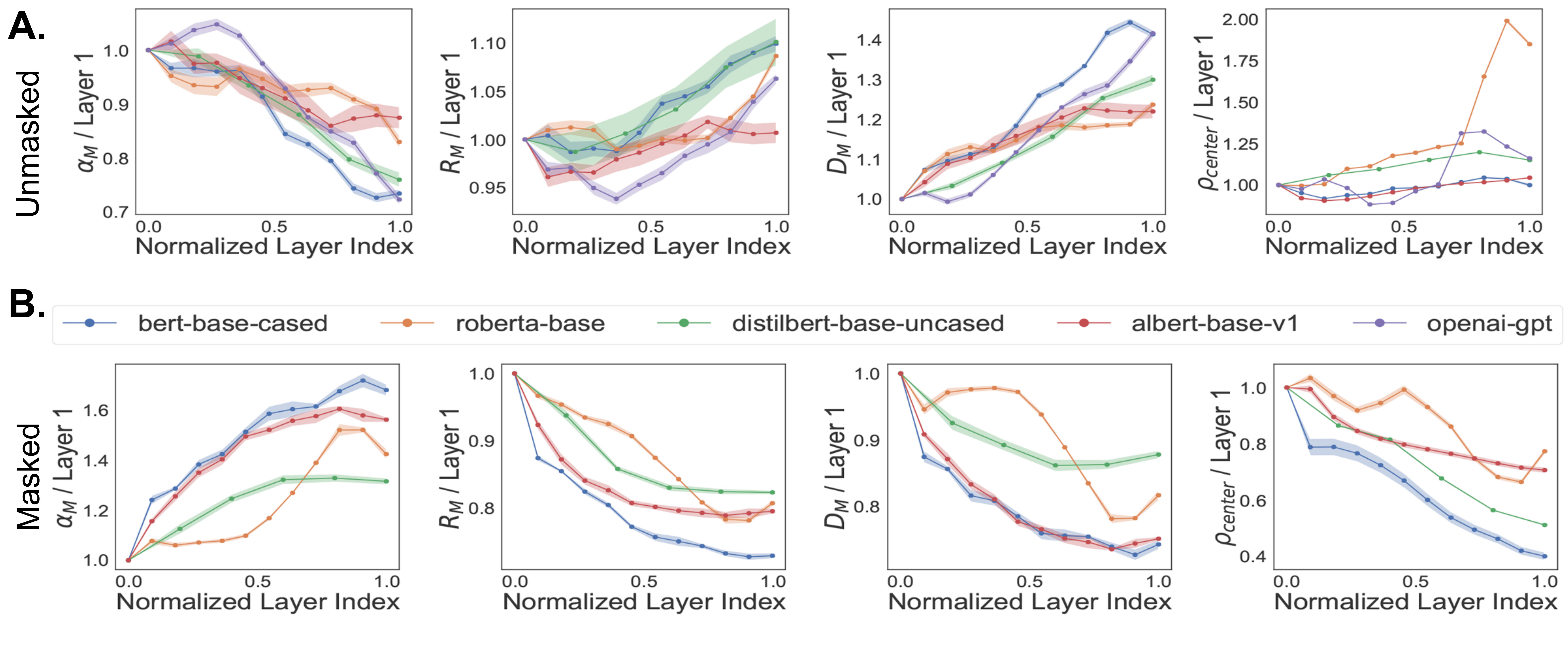}}
\caption{Geometric properties of semantic tags manifold in different models.} 
\label{fig:fig4_sem-tag}
\end{center}
\end{figure*}
For semantic tag manifolds, as reported in Figure~\ref{fig:fig4_sem-tag}, similarly to word manifolds and POS manifolds, the capacity decreases in the unmasked dataset and increases in the masked dataset. Similarly to POS tags, semantic tags also have high correlation with context-free word; as reported by \citet{bjerva_semtag}, the per-word most frequent class baseline for semantic tags has an accuracy of $77.39\%$. Therefore, in the masked case, since the model is trained to predict the word tokens which share information with the semantic tags, the manifold capacity increases. In the unmasked case, the inputs are word embedding vectors, which carry information about semantic tags, and the model tries to contextualize the inputs by their neighboring words. Contextualization can both entangle semantic tags manifold by decreasing the correlation between word tokens and their semantic tags and untangle semantic tags manifold by gaining information from neighbor words. These two competing effects lead to an overall decrease in manifold capacity, but this decrease has a much less magnitude than the decrease in word manifold capacity ($-0.6$ vs. $-0.06$). Manifold radius, dimension and center correlation also have similar trend as POS and word manifolds.

\paragraph{Named-entity Recognition}

\begin{figure*}[ht]
\begin{center}
\centerline{\includegraphics[width=2\columnwidth]{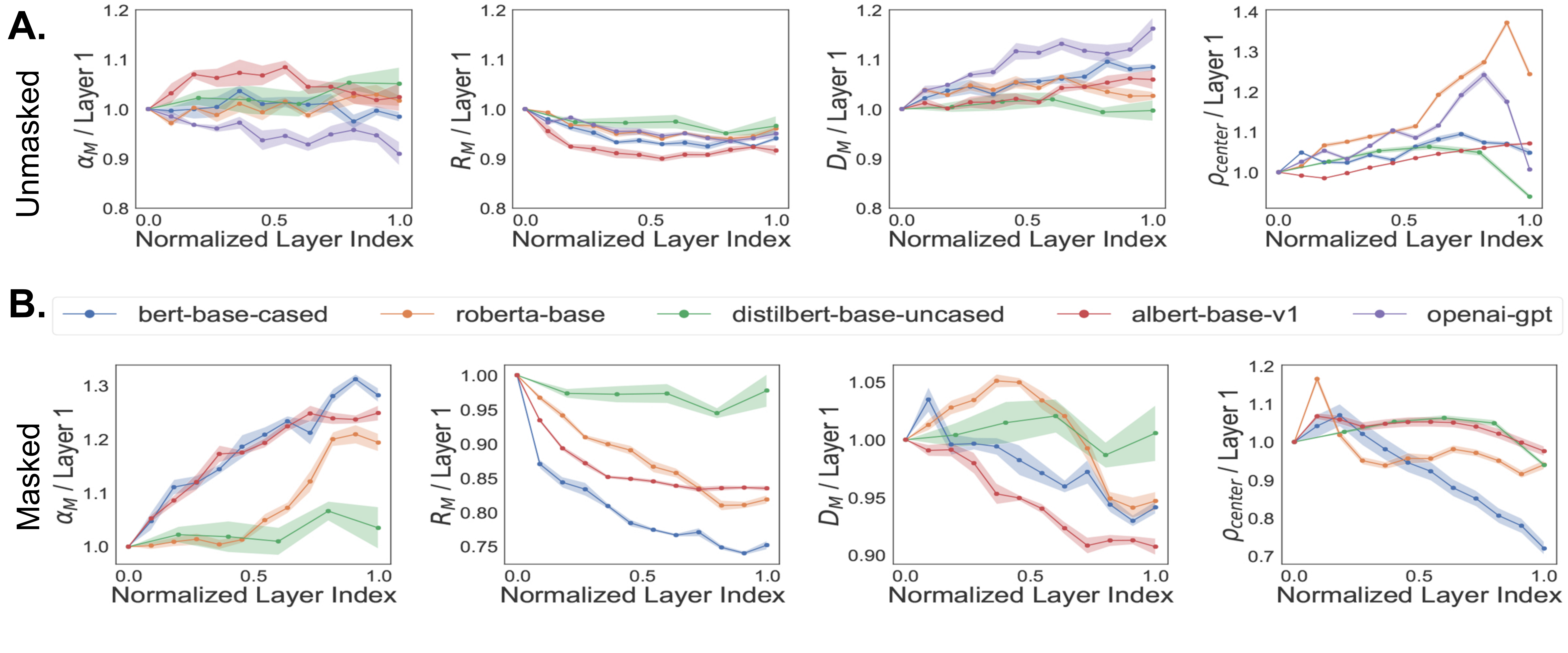}}
\caption{Geometric properties of NER manifold in different models.} 
\label{fig:fig4_ner}
\end{center}
\end{figure*}

For NER manifolds, as reported in Figure~\ref{fig:fig4_ner}, the different models express similar trend for both masked and unmasked data. 
For the unmasked data, the manifold capacity remains mostly unchanged across layers. This trend suggests a balance between losing information from correlation between words and NER label, and gaining information from contextualization by neighbor words. The geometric properties also show a competing effect between decreasing radius and increasing dimension and center correlation. For the masked data, the manifold capacity increases across layers (similar trend as word, POS and sem-tag). This trend is expected because the input tokens are masked and the model objective is to predict the masked word, which can carry some information about NER. Geometric properties show decreasing radius and center correlation, suggesting manifold untangling.  

\paragraph{Dependency Depth}
\begin{figure*}[ht]
\begin{center}
\centerline{\includegraphics[width=2\columnwidth]{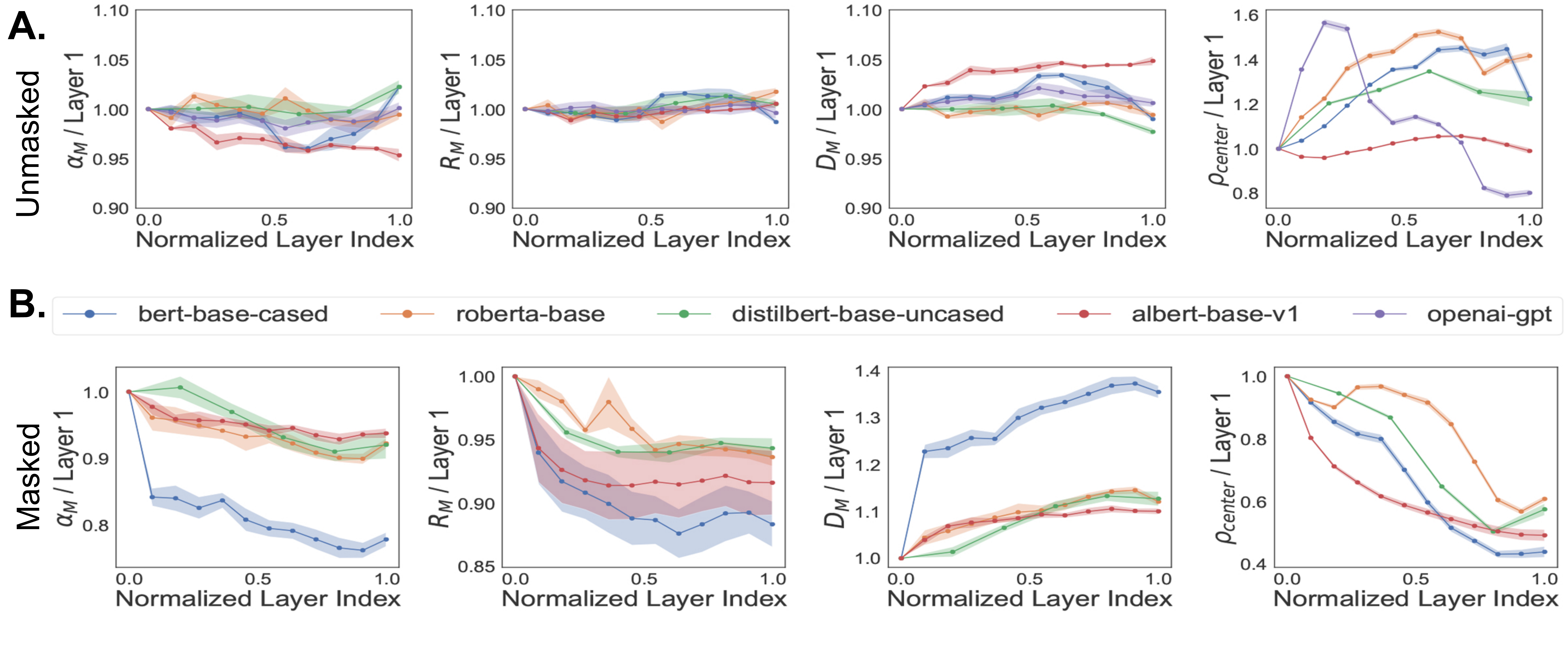}}
\caption{Geometric properties of dependency depth manifold in different models.} 
\label{fig:fig4_dep-depth}
\end{center}
\end{figure*}
For dependency depth manifolds, as reported in Figure~\ref{fig:fig4_dep-depth}, similar trend is observed for the different models in both masked and unmasked dataset. For unmasked data, the manifold capacity remains mostly unchanged. Manifold radius and dimension do not change significantly, while center correlation peaks at the intermediate layers. Since dependency depths are numerical values, higher center correlation may suggest a structured geometry relationship between different dependency depth clusters. \citet{hewitt2019} also reports similar results about syntactic parse tree peaks at the intermediate layers. For masked data, manifold capacity, radius and center correlation decreases across layers, while dimension increases. Generally, the manifold capacity and geometry measures for dependency depth manifolds are quite different from other manifolds. While other manifolds are {\it categorical} values, dependency depths are {\it numerical} values. A large capacity implies that category manifolds are well-separated for a classification task; however, since dependency depth manifolds have a numerical and transitive property, its geometry may not be optimized for classification capacity. Instead, dependency depth task may be explained better by a task that reflects such numerical and transitive properties such as a regression task, and the relation between the representation geometry and the regression performance will be explored as future work. 

\subsubsection{Correlation of Manifold Capacity and Task Performance in POS Fine-Tuned Model}
\label{SI:fine:tune}

\begin{table}[ht]
\centering
\begin{tabular}{c|c|c}
     update step & raw capacity  &  F1  \\ \hline
    1 & 0.0903 & 0.04 \\
    5 & 0.0915 & 0.11 \\
    10 & 0.0998 & 0.34 \\
    20 & 0.1362 & 0.55 \\
    50 & 0.2361 & 0.87 \\
\end{tabular}
\begin{tabular}{c|c}
     Pearson correlation & 0.9334 \\
\end{tabular}

\caption{Correlation of raw manifold capacity and F1 in POS fine-tuned model, unmasked data.}
\label{tab:F1:raw}
\end{table}  

\begin{table}[ht]
\centering
\begin{tabular}{c|c|c}
     update step & norm. capacity  &  F1  \\ \hline
    1 & 0.6111 & 0.04 \\
    5 & 0.6209 & 0.11 \\
    10 & 0.6839 & 0.34 \\
    20 & 0.9623 & 0.55 \\
    50 & 1.6274 & 0.87 \\
\end{tabular}
\begin{tabular}{c|c}
Pearson correlation & 0.9417 \\
\end{tabular}
\caption{Correlation of manifold capacity (normalized by embedding layer) and F1 in POS fine-tuned model, unmasked data.}
\label{tab:F1:norm}
\end{table}

When fine-tuning pre-trained {\tt bert-base-cased} model for POS task, a strong correlation is observed between the POS manifold capacity and F1 score across update steps for unmasked data, as reported in Table~\ref{tab:F1:raw} and Table~\ref{tab:F1:norm}. Specifically, Pearson correlation for raw capacity and F1 score and for normalized capacity and F1 score are $0.9334$ and $0.9417$ respectively. This result suggests that manifold capacity can capture task performance (F1 score) in POS task. Note that asked data is not shown because masked token is never seen during fine-tuning.
\end{document}